\documentclass[lettersize,journal]{IEEEtran}
\usepackage{cite}
\usepackage{amsmath,amssymb,amsfonts}
\usepackage{float}
\usepackage{graphicx}
\usepackage{epstopdf}
\usepackage{textcomp}
\usepackage{xcolor}
\usepackage{indentfirst}
\usepackage{tikz}
\usepackage{algorithm}  
\usepackage{algorithmicx}  
\usepackage{algpseudocode}
\usepackage{booktabs}
\usepackage{url}
\usepackage{bm}
\usepackage{subfigure} 
\usepackage{multirow}
\usepackage{makecell}
\usepackage{svg}

\ifCLASSINFOpdf
\else
\fi
\begin{document}
%
\title{Deep Domain Adaptation for Pavement Crack Detection}
%
%
%

\author{Huijun Liu,
        Chunhua Yang,
        Ao Li,
        Sheng Huang,~\IEEEmembership{Member,~IEEE,}
        Xin Feng,
        Zhimin Ruan,
        Yongxin Ge,~\IEEEmembership{Member,~IEEE}
\thanks{This work was partially supported by the National
	Natural Science Foundation of China (Grant no. 62176031), the Fundamental Research Funds
	for the Central Universities (Grant no. 2021CDJQY-018), and Humanities and Social Sciences of Chinese Ministry of Education Planning Fund (Grant no. 17YJCZH043). (Corresponding author: A. Li, Y. Ge.)
}
\thanks{H. Liu and C. Yang are with the College of Computer Science, Chongqing University, Chongqing 400044, China. 
}
\thanks{A. Li is with the School of Big Data and Software Engineering, Chongqing University, Chongqing 401331, China, and also with Chongqing City Management College, Chongqing 401331, China (email:li\_ao@cqu.edu.cn).}
\thanks{S. Huang and Y. Ge are with the Key Laboratory of Dependable Service Computing in Cyber Physical Society, Ministry of Education, Chongqing University, Chongqing 400044, China, and also with the School of Big Data and Software Engineering, Chongqing University, Chongqing 401331, China (email:yongxinge@cqu.edu.cn).}
\thanks{X. Feng is with the College of Computer Science And Engineering, Chongqing University of Technology, Chongqing 400054, China.}
\thanks{Z. Ruan is with China Merchants Road Information Technology (Chongqing) Co., Ltd, Chongqing 400060, China.}}

\maketitle

\begin{abstract}
 Deep learning-based pavement cracks detection methods often require large-scale labels with detailed crack location information to learn accurate predictions. In practice, however, crack locations are very difficult to be manually annotated due to various visual patterns of pavement crack. In this paper, we propose a Deep Domain Adaptation-based Crack Detection Network (DDACDN), which learns domain invariant features by taking advantage of the source domain knowledge to predict the multi-category crack location information in the target domain, where only image-level labels are available. Specifically, DDACDN first extracts crack features from both the source and target domain by a two-branch weights-shared backbone network. And in an effort to achieve the cross-domain adaptation, an intermediate domain is constructed by aggregating the three-scale features from the feature space of each domain to adapt the crack features from the source domain to the target domain. Finally, the network involves the knowledge of both domains and is trained to recognize and localize pavement cracks. To facilitate accurate training and validation for domain adaptation, we use two challenging pavement crack datasets CQU-BPDD and RDD2020. Furthermore, we construct a new large-scale Bituminous Pavement Multi-label Disease Dataset named CQU-BPMDD, which contains 38994 high-resolution pavement disease images to further evaluate the robustness of our model. Extensive experiments demonstrate that DDACDN outperforms state-of-the-art pavement crack detection methods in predicting the crack location on the target domain. 
\end{abstract}


%
\IEEEpeerreviewmaketitle

\section{Introduction}
\IEEEPARstart{C}{racks} are common pavement diseases that seriously threaten driving safety. In densely distributed pavement traffic, pavement cracks have become an important part of pavement detection and maintenance. In practice, due to various external forces, pavement cracks would present different types of visual patterns, such as longitudinal cracks, transverse cracks, alligator cracks, and potholes. Some of them are very small relative to the entire image, and some are hardline cracks on the rough ground. Hence, effectively detecting pavement cracks is quite challenging and has been studied for decades.
\begin{figure}[tbp]
	\centering
	\includegraphics[scale=0.6]{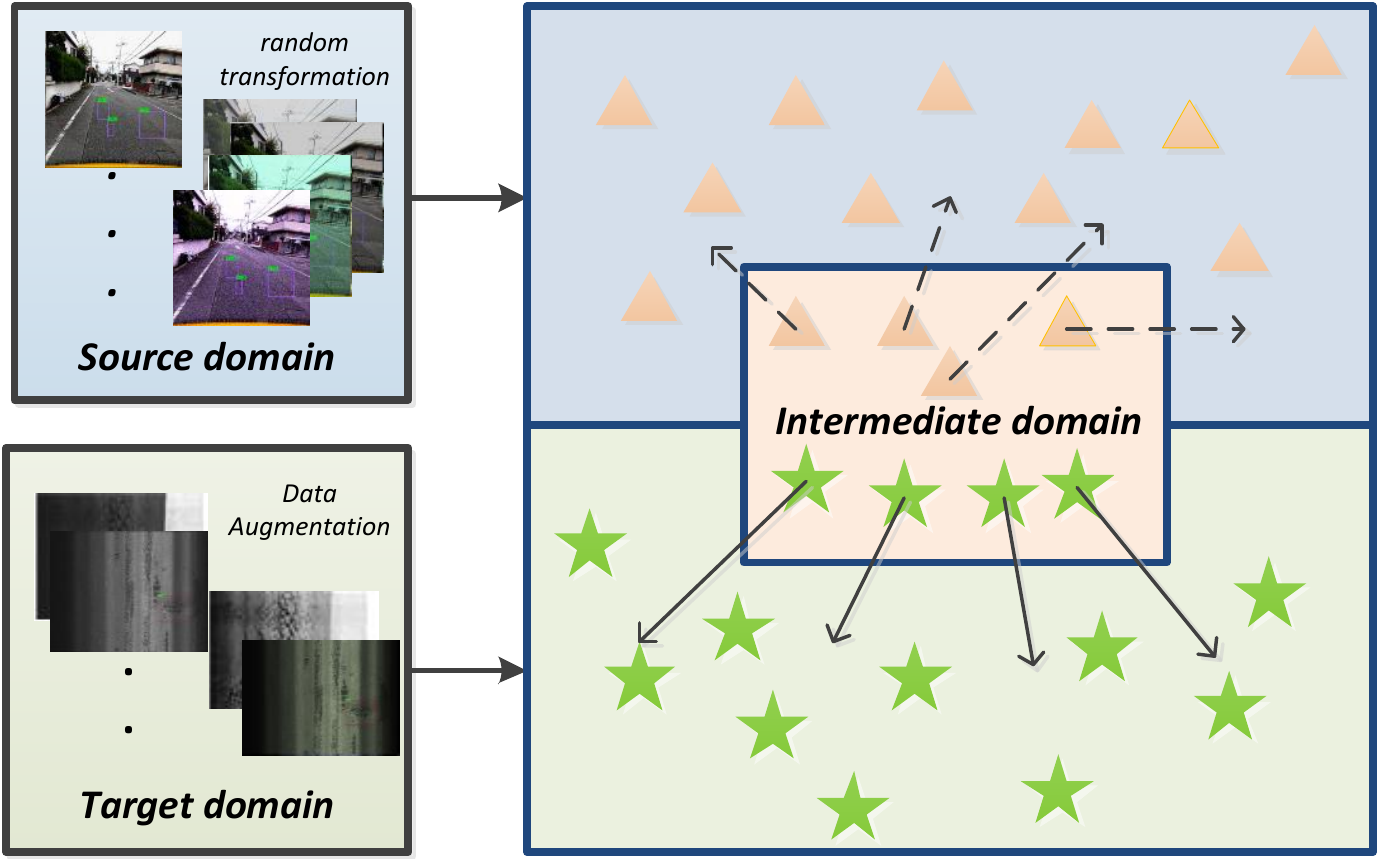}
	\caption{Illustration of the intermediate domain in DDACDN. We first enhance the target dataset in the method we proposed. Then, we utilize multiple methods to augment the source dataset and the target dataset. Finally, a new intermediate domain is constructed by features figigm both the augmented two domains, which is shown in the middle right part of the figure. Please refer to section \uppercase\expandafter{\romannumeral3} for specific details.}
	\label{fig:1}
\end{figure} 
Traditionally, pavement cracks are detected manually by pavement maintenance workers, which consumes a lot of manpower and resources, and has disadvantages of large personal subjective factors and low detection accuracy. Whereas automated detection will not only reduce detection cost, but also improve detection efficiency. However, pavement cracks are irregular and non-rigid objects, which have different sizes, shapes, and various discontinuous patterns. 
 All these problems increase the difficulty of crack detection, and the detected cracks are far from reaching the expected results. 

Current pavement crack recognition is mainly based on road images collected by CCD cameras mounted on vehicles or unmanned aerial vehicles using image processing technologies. The existing traditional crack detection methods can be divided into two categories: image processing-based methods and machine learning methods. Image processing-based methods \cite{amhaz2016automatic,fernandes2014pavement,shi2016automatic} mainly rely on hand-crafted and low-level features in pavement cracks; they enhance the continuity of the detected cracks. Machine learning method was introduced by Koch et al. \cite{koch2015review} for crack detection. \cite{kapela2015asphalt,cord2012automatic,ali2018vision,oliveira2008supervised} utilized low-level features extracted from road images to train classifiers, such as Support Vector Machine (SVM), Bayesian, and AdaBoost, greatly improving the robustness of crack detection. However, the low-level features for training only obtain shallow abstraction of cracks and cannot fully perceive the complex pavement surface.
 
 Different from traditional crack detection methods, recent deep learning-based methods learn different levels of pavement crack features for localization from annotated datasets, which can be roughly grouped into two categories: one-stage methods and two-stage methods. The one-stage methods \cite{yu2021real,bochkovskiy2020yolov4} usually predict crack category and location by regressing the features extracted in the network. The two-stage methods \cite{suh2018deep,ren2016faster,cha2018autonomous} usually generate region proposals, classify them in turn, and correct the position. These deep learning-based methods have achieved better performance than traditional crack detection methods. However, they usually require a large number of annotated data with detailed location information of pavement cracks. Vishal et al. \cite{mandal2018automated} classify more than seven categories of cracks by training the network on a location-labeled dataset. \cite{chen2019deep} proposes a multi-label neural network for road scene recognition and builds a large dataset of over 110,000 images. In \cite{majidifard2020pavement,asadi2021automated}, more than 7000 images are manually annotated with location information for training. However, obtaining such a large number of labels is quite labor-intensive and time-consuming. In addition, manual annotation may cause unexpected label deviation. Aiming at reducing the workload of manual annotation, a label augmentation method is proposed in \cite{zhang2020crackgan}, to generate partially accurate labels for training. In practice, however, the partially inaccurate labels are prone to interfere with the model training. Therefore, we propose a domain adaptation-based pavement crack detection method by annotating a small amount of data on the target domain. Domain adaptation \cite{wang2018deep} is a representative method in transfer learning. The source domain represents a domain that is different from the test set, and has rich crack location information annotations. The target domain represents the domain where the test samples have no location label or only a few labels. Moreover, most of the current pavement crack detection methods do not use multi-scale features of crack images for transfer learning, thus cannot transfer features of cracks with different scales from source domain to target domain well. Finally, to our best knowledge, previous works have not explored domain adaptation-based pavement crack detection methods to detect various pavement cracks, such as potholes and transverse, etc.
As far as this issue is concerned, we are the first to conduct multi-category crack detection through domain adaptation transfer with only image-level labels being available in the target domain.

In this study, to save the cost of labeling crack location information and solve crack disease detection problems on datasets without location annotation, we propose a Deep Domain Adaptation-based Crack Detection Network (DDACDN) for multi-category pavement crack detection, which effectively bridges domain discrepancy and obtains similar features on the two domains. Specifically, the proposed method converts the road image from the source domain, and the target domain to their corresponding feature space. A Multi-scale Domain Adaptation (MDA) strategy is designed to constrain the distance between multi-scale features from both domains by following the Multi-Kernel Maximum Mean Discrepancy (MK-MMD) method in \cite{long2015learning} to make the model adapt the feature patterns of cracks in the target domain. Furthermore, the two domain features are aggregated in an intermediate domain for further training to aggregate aligned domain features for domain invariant feature extraction, as shown in Fig. \ref{fig:1}. Finally, through the constraint training of source domain data and target domain data, the model obtains the discriminative knowledge in the target domain, and can learn the domain invariant features to effectively recognize cracks with few labeled data in the target domain. 
  In addition, to deal with the noise caused by the illumination of the target domain data, we design a simple image adaptive enhancement method.

The contributions of this paper can be summarized in the following aspects:
\begin{itemize}
	\item To the best of our knowledge, we are the first to apply deep domain adaptation idea in pavement crack detection, which provides a solution to the problem of crack detection without annotation of crack location.
	\item We propose a Deep Domain Adaptation-based pavement Crack Detection Network (DDACDN), which uses the knowledge of the source domain to perform detection on the target domain. Specifically, feature spaces that contain multi-scale features extracted from both the source and target domains are first aligned by using the proposed Multi-scale Domain Adaptation (MDA) strategy, and an intermediate domain is then constructed to aggregate the aligned domain features and facilitate the whole network for domain invariant feature extraction. Moreover, a simple yet effective image enhancement method is designed to effectively suppress the negative effects of illumination.
	\item We release a new, large-scale bituminous pavement disease dataset that is acquired from various real pavement scenarios. The dataset contains more than 38000 high-resolution pavement crack images and involves various types of diseases.
	\item We conduct extensive experiments and systematically compare our model with recent pavement detection methods, in terms of localization accuracy, robustness and cross-data generalization ability, etc. All experimental results demonstrate the superiority of our method in pavement crack detection.
\end{itemize}

\section {Related Work}
\subsection{Crack Detection Methods}
In the past decades, automatic pavement crack detection has been well researched. Generally, the proposed methods can be roughly divided into two streams: traditional crack detection methods and deep learning-based methods.

\textbf{Traditional crack detection method:} Intensity thresholding methods \cite{hu2010novel} is based on the difference between the grayscale value of the crack pixels and the surrounding pixels, which is particularly sensitive to the noise in the image. Hanzaei et al. \cite{hanzaei2017automatic} and Zhao et al. \cite{zhao2010improvement} use edge detection methods for crack detection, which utilize the jump change of the pixel value near the crack. However, it is shown only a set of disjoint crack fragments can be detected by these methods, which is not applicable in low-contrast images. 

Traditional methods can be combined with machine learning methods. For example, Quintana et al. \cite{quintana2015simplified} utilize support vector machines to train the classifier based on local feature descriptors, such as Local Binary Pattern (LBP), for crack recognition. Prasanna et al. \cite{prasanna2014automated} and Pan et al. \cite{pan2017object} adopt random forest (RF) for crack detection, which is a special bagging method that randomly extracts a part of features from features and finds the optimal solution among the extracted features. 
\cite{hoang2021computer} uses texture descriptors measured by color channel statistics, gray-level co-occurrence matrices, and local ternary patterns to extract texture information from pavement images to help SVM classification.
Moreover, the combination of three classifiers (SVM, RF, and artificial neural network) \cite{pan2018detection} has achieved good crack recognition results as well. These methods, however, usually extract feature vectors from sub-images, which only involve local receptive field and cannot accurately segment the complete cracks.

\textbf{Deep learning-based method:} As a part of machine learning, deep learning has the advantage of learning high-level semantic features. It has made great breakthroughs in the field of crack detection and classification. Zhang et al. \cite{2016Road} first propose a shallow neural network composed of four convolutional layers and two fully connected layers for patch-based image crack detection. In \cite{nhat2018automatic}, a convolutional neural network (CNN) is designed for crack detection, which includes a feature extraction network and a classification network. The first network is responsible for extracting features, and the second network is used for identification. In \cite{park2019patch}, a patch-based deep learning method is proposed for crack detection research on small datasets. \cite{yang2021structural} adopts YOLOv3 and ResNet18 to efficiently detect and classify crack images.
In \cite{gopalakrishnan2017deep,silva2018concrete}, the VGG16 \cite{simonyan2014very} backbone that has been pretrained on ImageNet \cite{5206848} dataset, is finetuned for crack classification. In \cite{lee2019robust}, a pixel-level network based on FCN is proposed, which combines the segmentation stage with the detection stage. In \cite{girshick2014rich}, R-CNN is proposed to perform object detection and obtain the final object position through bounding box regression, which is a two-stage algorithm. In \cite{ZHANG20208205,du2020pavement}, to better detect cracks, many improvements have been made based on the YOLO network, which is similar to the innovation made by \cite{gao2020high,feng2020pavement} based on Single Shot MultiBox Detector (SSD), and both of them are one-stage algorithms. 

These deep learning-based methods can achieve excellent performance in practical applications, but large-scale datasets annotated with detailed location information are required during the training process. In this paper, with the help of the knowledge transferred from the source domain, we attempt to utilize a few samples with location annotations for crack detection in the target domain.
\begin{figure*}[h]
	\centering
	\includegraphics[scale=0.65]{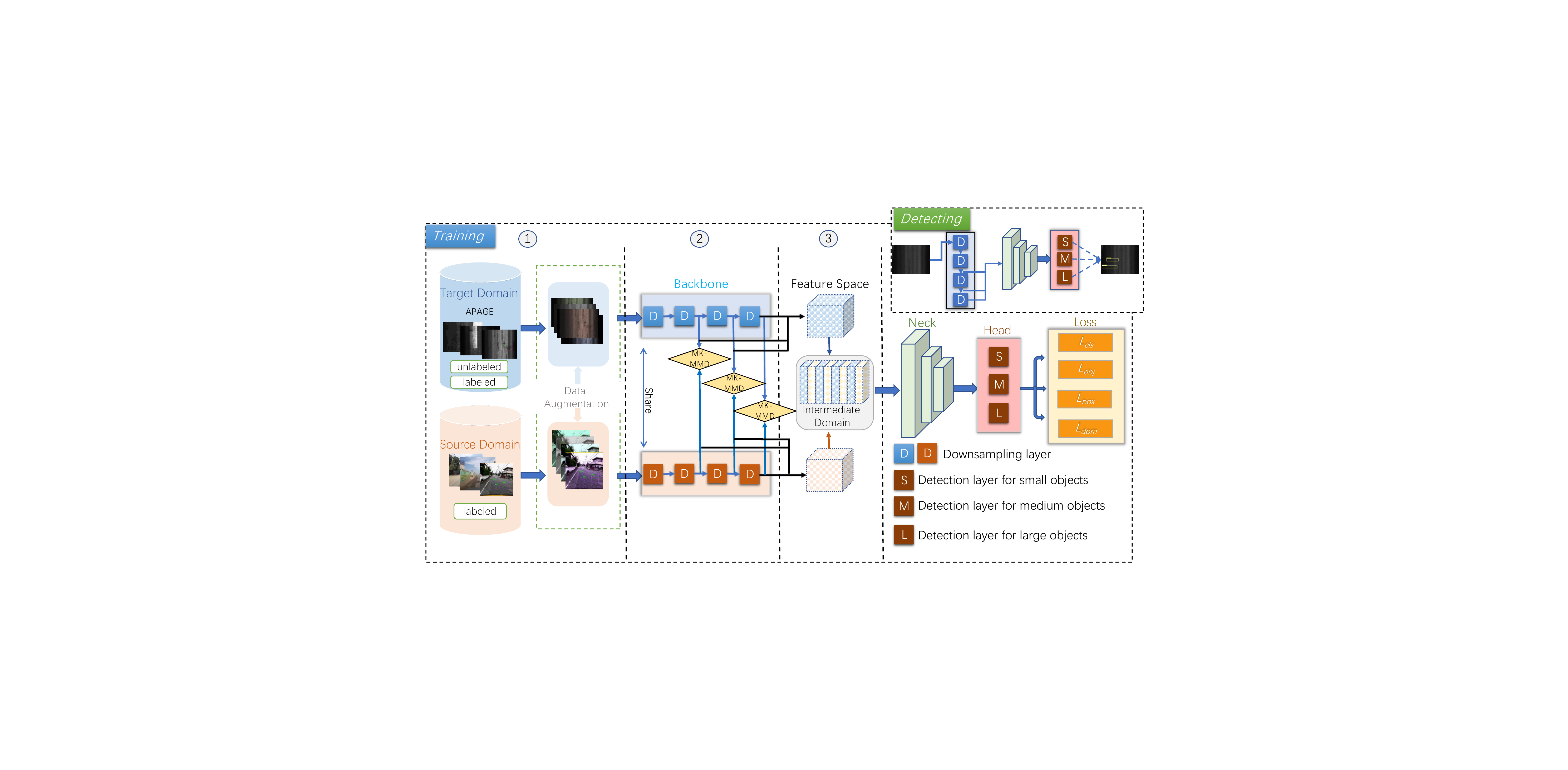}
	\caption{The architecture of the proposed method. Part one is the data preprocessing part, including the APAGE module and enhancement module. Part two is the feature extraction module. Part three is a domain adaptation mining module. We first send the data of the source domain and target domain into the network for training, and extract the features of both on three scales for domain adaptation. Then, we calculate the similarity of features in the two domains. Finally, an intermediate domain has been constructed to train the network.}
	\label{fig:2}
\end{figure*}
\subsection{Domain Adaptation}
Domain Adaptation (DA) \cite{ganin2015unsupervised,tzeng2017adversarial,wang2019transferable} is a representative method of transfer learning \cite{tan2018survey}, which aims to map the data of the source domain and target domain into a common feature space and make them as close as possible in the space. Therefore, to improve the accuracy on the target domain, the target function, which is trained on the source domain in the feature space, can be transferred to the target domain. In general, the data distributions of the source and target domains are different, and the source domain has a large amount of labeled data, while the target domain has no (or few) labeled data. 
DA-Faster-RCNN \cite{chen2018domain} designs domain adaptation components of two branches, image-level and instance-level, respectively, to alleviate the invariant domain difference at different annotation levels. By integrating an image-level multi-label classifier on the detection backbone, DA-Faster-ICR-CCR \cite{xu2020exploring} can obtain sparse yet critical image regions corresponding to the classification information. Meanwhile, at the instance level, it exploits the class consistency between image-level and instance-level predictions as a regularization factor to automatically find hard-aligned instances of the target domain.
In \cite{ma2019gcan}, an end-to-end graph convolutional adversarial network (GCAN) is proposed to achieve unsupervised domain adaptation through joint modeling data structures, domain labels, and class labels in a unified depth framework. In \cite{long2017conditional}, a principled framework for conditional adversarial DA is proposed, which is designed with two novel conditioning strategies: multilinear conditioning and entropy conditioning. The former captures the cross-covariance between feature representations and classifier predictions to improve the identifiability of the classifier, and the latter controls the uncertainty of classifier predictions to ensure transferability.

\section{Methodology}
\subsection{Problem Formulation and Overview}
Most existing deep learning-based crack detection methods use datasets with detailed annotations, but in most cases, we do not have such annotations and they are very difficult to obtain. On the contrary, image-level annotation is easy to obtain and has great value in practical applications. The method proposed in this paper carries out detection for pavement crack images in the target domain, where there is no crack location annotation, by taking advantage of the rich location information in the source domain.

Specifically, given a labeled source domain $D_s$ and some manually labeled data belonging to target domain $D_t$, we map them to the corresponding feature space respectively. To make the distribution more approximate, it is necessary to calculate the feature distances of different scales of the two domains respectively. Then, we construct the intermediate domain $D_i$ to further train the network to improve the ability to extract domain invariant features.
Fig. \ref{fig:2} illustrates the architecture of our method. We will give a detailed introduction in the next subsections.

Formally, we denote the domain adaptation-based pavement crack detection as follows: domain $D$ includes the feature space $X$ and the marginal probability distribution $P\left( X \right)$. $ X_s$ and $X_t$ represent the feature spaces of $D_s$ and $D_t$. For given $D_s$ and $D_t$, tasks on the two domains are the same. Task contains the category space Y and the conditional probability distribution $P\left( Y|X \right)$. $Y_s$ and $Y_t$ are used to represent the category space of $D_s$ and $D_t$ respectively. Source domain $D_s$ can be formulated as: $D_s=\left\{x_l,y_l \right\}_{l=1}^n$, where $y_l\in Y_s$ and $x_l\in X_s$. Target domain $D_t$ lacks annotated information, which can be formulated as: $D_t=\left\{x_u\right\}_{u=1}^{r}$.
\subsection{Adaptive Patch Augmentation and Global Equalization}
Since the pavement images of $D_t$ are captured from different road conditions and at different times, cracks in the pavement images usually present seriously uneven illumination and noise. As a preprocessing method, CNN will have difficulties such as insufficient datasets and missing data annotation. And because the dataset cannot cover all the variations of road conditions, there is a bottleneck of using CNN in the performance of preprocessing. Conventional linear transformations can only brighten or darken the whole image and do not improve the overall mean square error of the image, while nonlinear transformations will enhance or reduce the region of extreme gray value, but cannot achieve the overall enhancement in the entire gray value interval.

In order to enhance balanced pavement images, we propose a simple and effective preprocessing method called Adaptive Patch Augmentation and Global Equalization (APAGE), which is based on Gamma Correction and CLAHE method proposed in \cite{109340}.
APAGE is based on the traditional digital image method, which relies on manual processing experience, and has relatively low data requirements. At the same time, this manual compensation can take as many road variations as possible into account. Specifically, APAGE can be divided into the following two steps:
\subsubsection{Patch Augmentation} The input image is divided into $R_{row} \times R_{col}$ small patches, and each patch has a size of $100 \times 100$ pixels. Then, for each patch, we use Gamma correction to enhance the dark part of each patch, where the Gamma $\Gamma$ is set from 0.5 to 2, and the incremental step size is 0.1. Finally, the $\Gamma$ that makes the patch has the largest variance is used as the exponent.
\subsubsection{Global Equalization} When all patches are transformed, the gray value transition between adjacent patches may be discontinuous. Therefore, CLAHE is used to equalize the overall image to suppress the negative effects of discontinuous gray value transitions.

In our experiment, to make full use of each pixel of the image, we empirically design the image blocking strategy and the patch size according to the resolution size of the pavement image.

\subsection{Domain Adaptation Mining}
The pavement images from the target domain $D_t$ is firstly enhanced by using APAGE. Then, we augment it together with the data of $D_s$ and input them into the backbone network for feature extraction, respectively. To learn the essential crack features, we intentionally make the two backbone network with shared weights. 

To solve the problem of less labeled data in the target domain, we should transfer the discriminative knowledge in the source domain to the target domain for recognition. Moreover, selecting suitable hidden layers is very important to calculate the domain loss, which can obtain similar features extracted by the model on the two domains. In this study, we adopt the Multi-scale Domain Adaptation (MDA) strategy to extract the features on the two domains from three scales, as shown in the second part of Fig. \ref{fig:2}. Hence, the fine-grained crack discriminative features can be learned to solve the problem that some cracks account for extremely small proportion in the image. Specifically, MK-MMD is introduced to calculate the domain loss from the three-scale features, namely, the 4-th, 6-th, and 9-th hidden layers of the backbone network, that is, the second, third, and fourth downsampling layers in Fig. \ref{fig:2}. Experiments have proven that the lower-level
layers can well represent the edge information of cracks. Also, the deeper the convolutional layer of the network, the more domain private domain features are extracted. We use MK-MMD to combine high-level and low-level domain losses to train the network. MK-MMD assumes that the optimal kernel can be obtained by the linear combination of multiple kernels, and its squared formulation is defined as:
\begin{equation}
	d_k^2(D_s,D_t)=\left \Arrowvert \frac{1}{N}\sum_{i=1}^{n}\phi(x_l)-\frac{1}{R}\sum_{j=1}^{r}\phi(x_u) \right \Arrowvert  _{H_k}^2  ,\label{mkmmd}
\end{equation}
where $H_k$ denotes the reproducing kernel Hilbert space endowed with a characteristic kernel $k$. The feature kernel, $k(x_l,x_u)=\langle \phi(x_l),\phi(x_u) \rangle $, associated with the feature map $\phi$, is defined as the convex combination of $m$ positive-definite kernel. 
The advantages of using multiple kernel functions to calculate the correlation and combining the crack features at three different scales are as follows. First, it essentially bridges the domain discrepancy underlying the conditional distribution. Second, it avoids the problem that a single kernel function may cause poor MMD with different domain distributions of crack features in Hilbert space.

\begin{algorithm}[]
	\caption{DDACDN-based Pavement Crack Detection} 
	{$\textbf{Input}$: Labled source domain $D_s$, A few labeled target domain $D_t$}
	\label{alg:1}
	
	{$\textbf{Output}$: A discriminative backbone model}
	\begin{algorithmic}[1]
		\State  $\textbf{Initialise}$: Model parameters; $\theta$ 
		\For{i=1$\rightarrow$Max Iteration } 
		\For{c=1$\rightarrow$$N_c$}
		\State Randomly selecting some data $d_t^c=\{x_u|x_u\in D_t,u=1,2,\cdots,M_c \}$;
		\State Utilizing APAGE to suppress negative effects of illumination in $D_t$;
		\EndFor
		\State Augmenting images from $d_t=\{d_t^c|c\in N_c\}$ and $D_s$; 
		\State Inputing $d_t$ and the data of $D_s$ into the network to extract features;
		\State Calculating the MK-MMD of $d_t$ and $D_s$ on the three feature scales and using it as the domain loss;
		\State Using these features from $d_t$ and $D_s$ to construct intermediate domain $D_i$ to continue training the network;
		\State Updating the network parameters $\theta$ through the overall loss function $\mathcal{L}$  in Equation \ref{ovreallloss};
		\EndFor
		\State \Return $\theta=\theta^\ast$
	\end{algorithmic}
\end{algorithm}

For feature space and category space, equations $X_i\subset X_s, X_i\subset X_t$ and $Y_i\subset Y_s, Y_i\subset Y_t$ are true,
	where $X_i$ and $ Y_i$ represent the feature space and category space of $D_i$ respectively. Due to $X_s\ne X_t$ and $Y_s\approx Y_t$, we apply the MDA strategy to each batch of data and use the domain losses, which are obtained by the feature extraction module on the three scales, to make $X_s$ and $X_t$ be closer. Furthermore, to aggregate aligned domain features for domain invariant feature extraction, we construct a new intermediate domain through the feature space on the two domains to aggregate features for further training. The intermediate domain contains the features of the two domains due to the multi-scale domain alignment by MDA; thus, we can obtain $X_i\subset (X_s\cap X_t$) and $Y_i\approx Y_s\approx Y_t$. Algorithm \ref{alg:1} presents the specific steps of our method, where $N_c$ represents the total number of categories, and $M_c$ represents the total number of data required for each category.

\begin{figure}[h]
	\centering
	\includegraphics[scale=0.75]{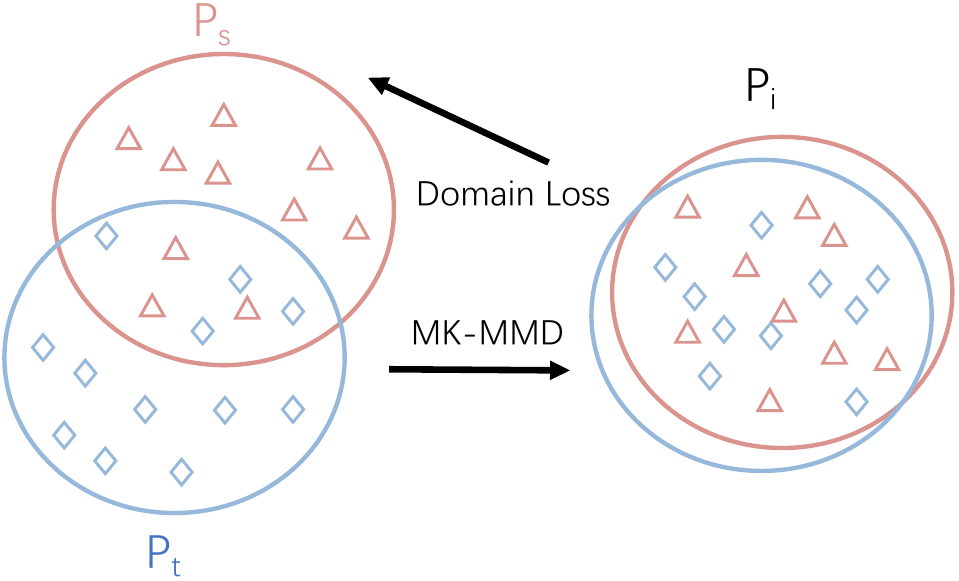}
	\caption{Process of domain adaptation. Although the conditional probability distributions of $D_s$ and $D_t$ are different in general, there are still some similarities, because the category spaces are similar. The network is constantly trained by calculating the domain loss by MK-MMD, and the domain invariant features of two domains can be extracted. Therefore, the conditional probability distributions on the two domains are aligned, as shown by $P_i$.}
	\label{fig:3}
\end{figure}

The Logical relation of conditional probability distributions $P_s\left( Y_s|X_s \right)$, $P_t\left( Y_t|X_t \right)$, and $P_i\left( Y_i|X_i \right)$ can be described in Fig. \ref{fig:3}. Although $P_s\left( Y_s|X_s \right)$ and $P_t\left( Y_t|X_t \right)$ are different, there is an intersection between the two domains because the data types in $D_s$ have corresponding types in $D_t$. We randomly select some data from $D_t$ and construct $D_i$ together with $D_s$. Therefore, $D_i$ contains the feature space and category space on both domains, and the domain loss calculated by MK-MMD is used to iteratively train the network. The domain invariant features extracted by the network are more abundant than before, and it can be approximated that $D_i$ aligns the conditional probability distribution of $D_s$ and $D_t$.

\subsection{Loss Function}
The loss function $\mathcal{L}$ of our model contains 4 components: classification loss $\mathcal{L}_{cls}$, confidence loss $\mathcal{L}_{obj}$, bounding box loss $\mathcal{L}_{box}$ and domain loss $\mathcal{L}_{dom}$. To better detect small crack objects, we start end-to-end training from $S=3$ scales,
and each scale divides the input image into $S_i\times S_i$ grids. Furthermore, each grid is responsible for predicting $M$ bounding boxes.

\subsubsection{$\mathcal{L}_{cls}$ and $\mathcal{L}_{obj}$}
 $\mathcal{L}_{cls}$ and $\mathcal{L}_{obj}$ are calculated by Focal Loss (FL) \cite{lin2017focal}, which is modified on the basis of the binary cross-entropy loss with logits,
\begin{equation}
	\begin{split}
	\mathcal{L}_{bcel}(p,\hat{p})=-\left\{p \log (\sigma (\hat{p}))\right. \\
\left.	+(1 - p) \log (1-\sigma (\hat{p})\right\}. \label{bce}
\end{split}
\end{equation}
The $\hat{p}$ and $p$ denote the predictive value and target value. $\sigma(\hat{p})$ is a sigmod function.
Therefore, 
FL can be expressed as:
\begin{equation}
	\begin{aligned}
		\mathcal{F}_L(p_j^c,&\hat{p}_j^c) =\alpha\{1-[(1-p_j^c)(1-\sigma(\hat{p}_j^c))\\ 
& +p_j^c\sigma(\hat{p}_j^c) ]\}^\gamma \ast\mathcal{L}_{bcel}(p_j^c,\hat{p}_j^c),
	\end{aligned}
\end{equation}
where $p_j^c$ and $\hat{p}_j^c$ indicate the target label and the results of model predictions for the $c$ category of the $j$-th grid, respectively. In this paper, we set $\gamma$ to 1.5 throughout all the experiments.
 $\alpha$ represents the category weight,
 which is related to $p_j^c$ and is initialized to 0.25. Therefore, $\mathcal{L}_{cls}$ can be mathematically denoted as follows:
\begin{equation}
	\mathcal{L}_{cls}=\sum_{i=1}^3\sum_{j=1}^{S_i^2}\mathbb{I}_j^{obj}\sum_{c\in classes}^{}\mathcal{F}_L(p_j^c,\hat{p}_j^c),
\end{equation}
where $\mathbb{I}_j^{obj}=1$ indicate that the $j$-th grid is responsible for a real object $obj$, otherwise $\mathbb{I}_j^{obj}=0$. Due to the fact that $\mathcal{L}_{obj}$ uses the same loss as $\mathcal{L}_{cls}$, we can express it as:
\begin{equation}
	\mathcal{L}_{obj}=\sum_{i=1}^3\mathcal{H}_i\sum_{j=1}^{S_i^2}\sum_{m=1}^{M}\mathbb{I}_{jm}^{obj}\mathcal{F}_L(C_{j}^m,\hat{C}_{j}^m),
\end{equation}
where $C_{j}^m$ and $\hat{C}_{j}^m$ indicate the true confidence and predictive confidence of the $m$-th bounding box of the $j$-th grid respectively, and $\mathcal{H}_i$ is the balance parameter of the $i$-th scale. 
\subsubsection{$\mathcal{L}_{box}$}
GIoU loss \cite{rezatofighi2019generalized} is used to calculate $\mathcal{L}_{box}$, which sets IoU \cite{yu2016unitbox} directly as the regression loss and can be expressed in:
\begin{equation}
	\mathcal{L}_{giou}=\frac{B_{jm}^t\bigcap B_{jm}^p}{B_{jm}^t\bigcup B_{jm}^p}-\frac{B_{jm}^c-(B_{jm}^t\bigcup B_{jm}^p)}{B_{jm}^c}, 
\end{equation}
where $B_{jm}^t$ and $B_{jm}^P$ are the target box and the prediction box of the $m$-th bounding box of the $j$-th grid respectively. $B_{jm}^c$ is the minimum bounding box in which $B_{jm}^t$ and $B_{jm}^P$ can be wrapped. Therefore, $\mathcal{L}_{box}$ can be denoted as follows:
\begin{equation}
	\mathcal{L}_{box}=\sum_{i=1}^3\sum_{j=1}^{S_i^2}\sum_{m=1}^{M}\mathbb{I}_{jm}^{obj}(1-\overline{\mathcal{L}}_{giou}).
\end{equation}
\subsubsection{$\mathcal{L}_{dom}$}
$\mathcal{L}_{dom}$ is calculated using MK-MMD, which is shown in equation \ref{mkmmd}. We think that domain adaptation on three scales can well maximize domain confusion, which can be expressed as:
\begin{equation}
	\mathcal{L}_{dom}=\sum_{i=1}^3 \beta_i d_k^2(D_s,D_t),
\end{equation}
where $\beta_i$ represents the weight coefficients of the domain loss in the three scales.
Hence, we get the total loss function $\mathcal{L}$, which can be expressed as:
\begin{equation}
	\mathcal{L}=\eta_{box}\mathcal{L}_{box}+\lambda_{obj}(\eta_{cls}\mathcal{L}_{cls}+\eta_{obj}\mathcal{L}_{obj})+\mathcal{L}_{dom},   \label{ovreallloss}
\end{equation}
where $\eta_{box}$, $\eta_{cls}$ and $\eta_{obj}$ represent their weight in $\mathcal{L}$ respectively. $\lambda_{obj}$ indicates whether the batch size has an object, if there is an object, $\lambda_{obj}=1$, otherwise $\lambda_{obj}=0$.

\section{Experiment}
\subsection{Datasets and Setup}
We evaluate the proposed method on two datasets, CQU-BPDD \cite{9447759} and RDD2020 \cite{arya2021rdd2020}. A commonly used pavement crack segmentation dataset, namely CFD \cite{shi2016automatic}, is used to validate the cross-data generalization ability of DDACDN. However, CFD is a small dataset with only 118 crack images and the cracks are mainly large cracks. \begin{figure}[h]
	\centering
	\includegraphics[scale=0.7]{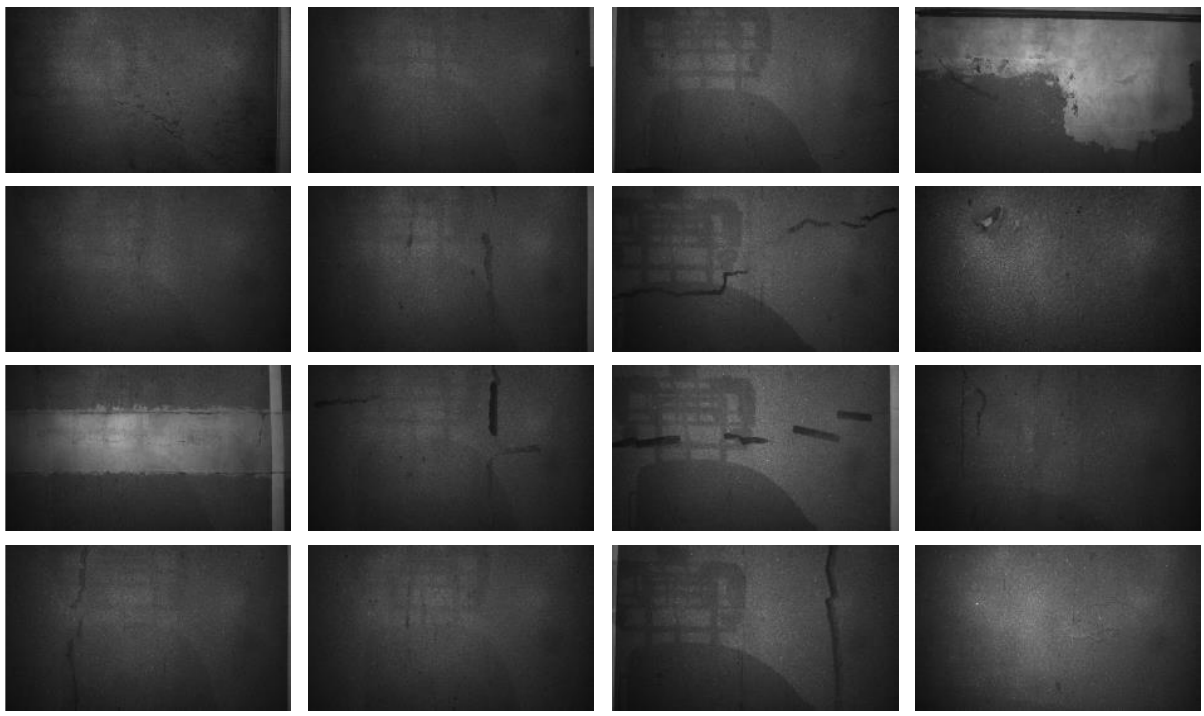}
	\caption{Some examples in CQU-BPMDD dataset which involves various pavement distress such as transverse crack, longitudinal crack, massive crack, repair, loose, wave crowding, pothole.}
	\label{fig:4}
\end{figure} Therefore, to verify the cross-data generalization of DDACDN more comprehensively, we construct a new large-scale Bituminous Pavement Multi-label Disease Dataset called CQU-BPMDD. \textbf{F$_1$}-score, \textbf{P}recision, \textbf{R}ecall, and \textbf{A}ccuracy are used for evaluation.
\begin{figure}[h]
	\centering
	\includegraphics[scale=0.45]{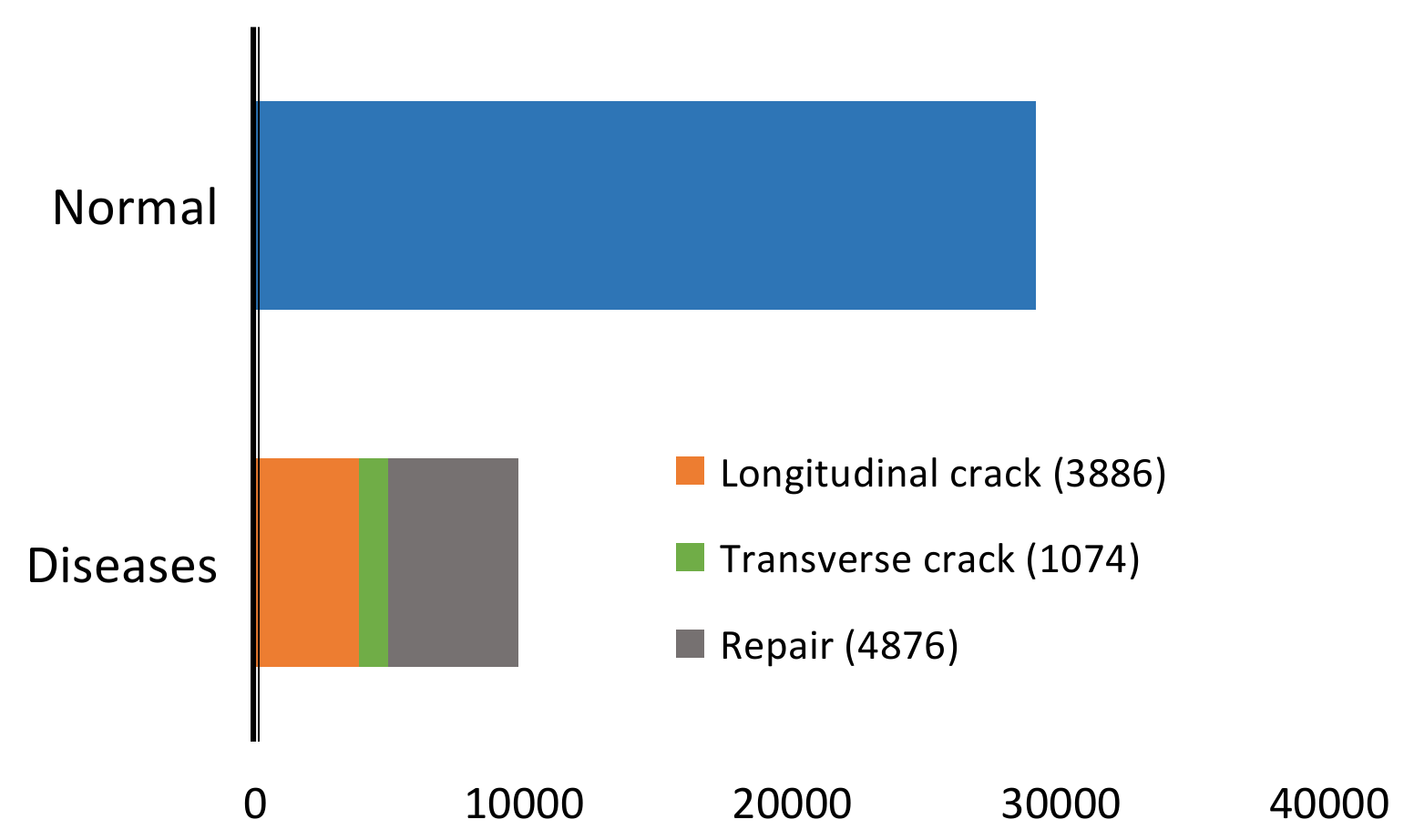}
	\caption{The sample distribution of CQU-BPMDD dataset, which is provided by China Merchants Road Information Technology (Chongqing) Co.,Ltd and Chongqing university. It contains 29143 normal images and 9851 diseases images. Note that the other four disease categories are not shown due to the current number of them is too less.}
	\label{fig:5}
\end{figure}
\begin{table*}[h]
		\caption{The comparison of different advanced pavement crack
			detection models on CQU-BPDD dataset. Res50 and Res101 represent ResNet50 and ResNet101, respectively.}
		\label{tab:1}
		\renewcommand\arraystretch{1.5}
		\centering
		\setlength{\tabcolsep}{1.6mm}{
			\begin{tabular}[htb]{p{3.2cm}|| c| c| c|| c| c| c||c |c |c||c| c| c||c}\Xhline{1.2pt}
				\multirow{2}{*}{Methods}&\multicolumn{3}{c||}{Longitudinal crack}&\multicolumn{3}{c||}{Transverse crack}&\multicolumn{3}{c||}{Alligator crack}&\multicolumn{3}{c||}{Pothole}&\multirow{2}{*}{Acc}\\
				\cline{2-13} 
				&P&R&F$_1$&P&R&F$_1$&P&R&F$_1$&P&R&F$_1$&\\
				\Xhline{1.2pt}
				Faster-RCNN (Res50) \cite{ren2016faster}&67.2\%&33.8\%&44.9\%&92.2\%&27.4\%&42.2\%&68.4\%&90.1\%&77.8\%&75.3\%&41.7\%&53.7\%&59.3\%\\
				Faster-RCNN(VGG16) \cite{ren2016faster}& 77.7\%&52.9\%&62.9\% & 90.7\%&58.3\%&71.0\%&90.5\%&83.3\%&86.8\%&59.6\%&47.4\%&52.8\%&66.2\%\\
				Faster-RCNN (Res101) \cite{ren2016faster}& 86.5\%&53.0\%&65.7\% & 92.4\%&66.0\%&77.1\%&90.8\%&86.6\%&88.6\%&55.9\%&45.3\%&50.1\%&68.9\%\\
				RFCN \cite{dai2016r}& 73.6\%&65.3\%&69.2\% & 91.3\%&54.6\%&68.4\%&91.3\%&84.9\%&88.0\%&59.9\%&67.1\%&63.3\%&72.8\%\\
				FPN \cite{lin2017feature}& 87.2\%&62.0\%&72.5\% & 93.6\%&73.8\%&82.5\%&88.9\%&91.0\%&89.9\%&54.9\%&43.5\%&48.5\%&74.2\%\\
				Cascade-FPN \cite{cai2018cascade} & 92.7\%&63.0\%&75.1\% & 95.3\%&75.5\%&84.3\%&95.5\%&88.9\%&92.1\%&76.1\%&56.1\%&64.5\%&75.5\%\\
				DA-Faster-RCNN\cite{chen2018domain}& 95.9\%&68.8\%&80.1\% & 96.9\%&78.7\%&86.9\%&99.3\%&86.0\%&92.2\%&90.2\%&42.6\%&57.8\%&74.5\%\\
				DA-Faster-ICR-CCR \cite{xu2020exploring}& 96.5\%&74.5\%&84.1\% & 96.3\%&82.1\%&88.6\%&99.3\%&88.2\%&93.5\%&90.9\%&45.9\%&61.0\%&77.9\%\\
				YOLOv4 \cite{bochkovskiy2020yolov4}& 89.5\%&68.6\%&77.7\% & 87.8\%&63.7\%&73.8\%&95.1\%&82.0\%&88.0\%&83.0\%&39.6\%&53.6\%&75.8\%\\
				Baseline & 96.4\%&70.6\%&81.5\% & 97.4\%&80.1\%&87.9\%&99.5\%&88.2\%&93.5\%&87.7\%&48.0\%&62.1\%&76.9\%\\
				\hline
				DDACDN (Ours) & \textbf{97.5\%}&\textbf{78.3\%}&\textbf{86.9\%} & \textbf{97.5}\%&\textbf{86.3\%}&\textbf{91.5\%}&\textbf{99.6\%}&\textbf{91.9\%}&\textbf{95.6\%}&\textbf{93.7\%}&55.8\%&\textbf{70.0\%}&\textbf{82.6\%}\\
				\Xhline{1.2pt}
			\end{tabular}
	}
\end{table*}

	\begin{table}[t]
	\caption{The category correspondence between RDD2020 and CQU-BPDD.}
	\label{tab:2}
	\renewcommand\arraystretch{1.5}
	\centering
	\setlength{\tabcolsep}{8mm}{
		\begin{tabular}{ c  c }\Xhline{1.2pt}  
			RDD2020 & CQU-BPDD  \\
			\hline	
			
			Longitudinal crack & Longitudinal crack  \\
			
			Transverse crack & Transverse crack \\
			
			Alligator crack & Crack and Massive crack \\
			
			Pothole  &   Loose \\
			\Xhline{1.2pt}
	\end{tabular}}
\end{table}

\subsubsection{Datasets}
	In our method, CQU-BPDD and RDD2020 always represent $D_t$ and $D_s$ respectively. CQU-BPDD contains 60059 crack images of bituminous pavement and includes seven kinds of pavement diseases. The training set has 49919 images, and the test set has 10140 images. Each pavement image corresponds to a 2$\times$3 meters pavement patch of highways with a resolution of 1200$\times$900, which has only image-level labels. RDD2020 consists of 26620 pavement images collected from Japan, India, and the Czech Republic, including a training set with a size of 21041, and there are four pavement disease categories. The resolution of each image in the training set is 600$\times$600. 
	
	CFD initially consists of 118 images with a resolution of 480$\times$320. Since CFD dataset is designed to study the so-called road crack detection task which is essentially a road crack segmentation task from the perspective of computer vision, and all samples are actually diseased images.

	 The common characteristic of the aforementioned three datasets is that the proportion of cracks in the image is relatively large, however, in practice, the crack targets usually look very small in the captured image. Hence, RDD2020 and CQU-BPDD are used as the source domain and target domain of our experiment, respectively, while CFD is used for cross-data generalization, which is not comprehensive enough. In order to better reflect the authenticity of the experiment, we construct a new dataset called CQU-BPMDD\footnote {The Dataset website: https://github.com/ychxff/CQU-BPMDD}, which is provided by China Merchants Road Information Technology (Chongqing) Co.,Ltd and Chongqing University. CQU-BPMDD contains 9851 diseases images and 29143 normal images, collected from highways at different regions in southwestern China. There are seven categories of diseases images, including longitudinal crack, transverse crack, repair, loose, pothole, massive crack, and wave crowding, as shown in Fig. \ref{fig:4}. Different from CQU-BPDD, CQU-BPMDD is a crack disease dataset with all images are imaged using exposure compensation, and most of them are small crack images, which are more in line with the actual scene and difficult to detect. Cross-Dataset validation on this dataset can better reflect cross-dataset generalization ability of our model. The resolution of each image is 3692$\times$2147, and the data distribution is shown in Fig. \ref{fig:5}. The usage of this dataset is described in the Cross-Dataset Validation section.

\subsubsection{Evaluation Metrics}
We conducted experiments on the above datasets. For each image, we calculate the \textbf{P}recision, \textbf{R}ecall, \textbf{F$_1$}-score, and \textbf{Acc}uracy by comparing the detected results with the true labels. \textbf{F$_1$}-score can reflect the overall index of performance evaluation. \textbf{Acc}uracy can reflect the ability of the model to detect four categories. These values are calculated based on $TP$, $TN$, $FP$, and $FN$, which can be mathematically represented as follows,
\begin{equation}
  \textbf{P}=\frac{TP}{TP+FP},\quad \textbf{R}=\frac{TP}{TP+FN}, \nonumber
\end{equation}
\begin{equation}
  \textbf{F}_1=\frac{2\times P\times R}{P+R}, \quad \textbf{Acc}=\frac{TP+TN}{TP+TN+FP+FN}, \nonumber
\end{equation}
where $TP$, $FP$, $FN$, and $TN$ indicate the numbers of true positives,false positives, false negatives, and true negatives respectively.
	\begin{figure*}[h]
	\centering
	\subfigure[Longitudinal crack]{\includegraphics[scale=0.27]{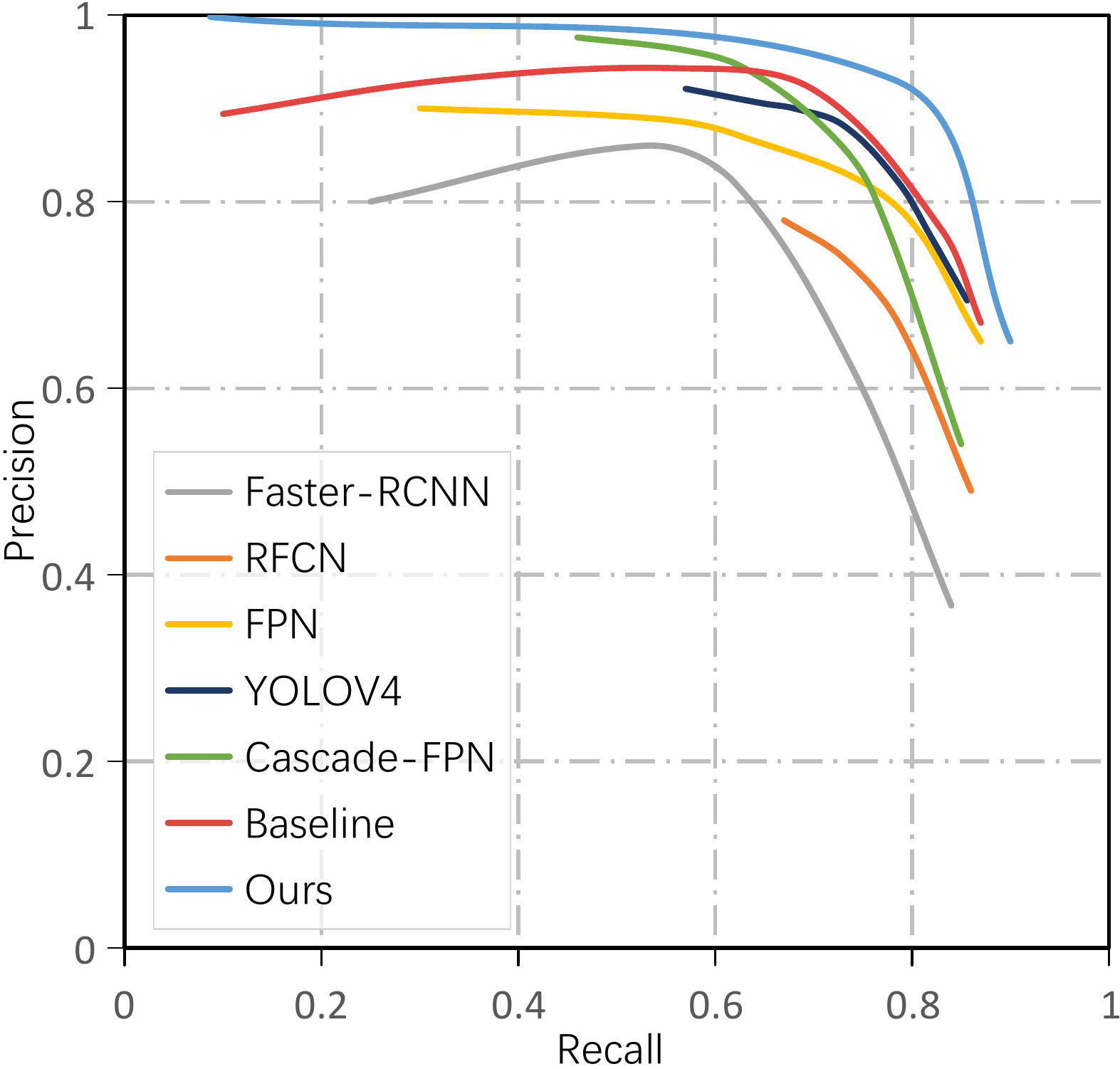}}
	\subfigure[Transverse crack]{\includegraphics[scale=0.27]{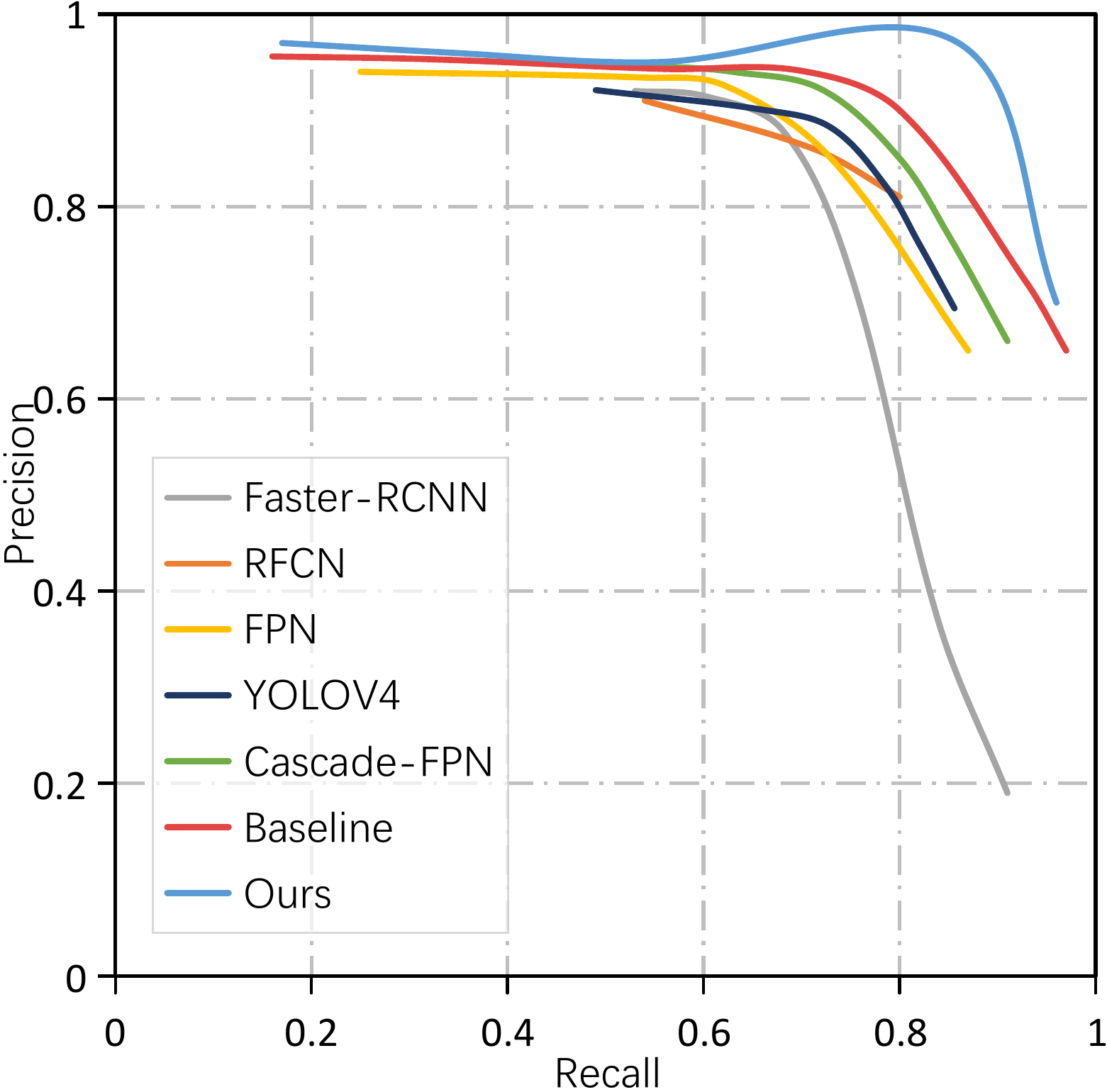}}
	\subfigure[Alligator crack]{\includegraphics[scale=0.27]{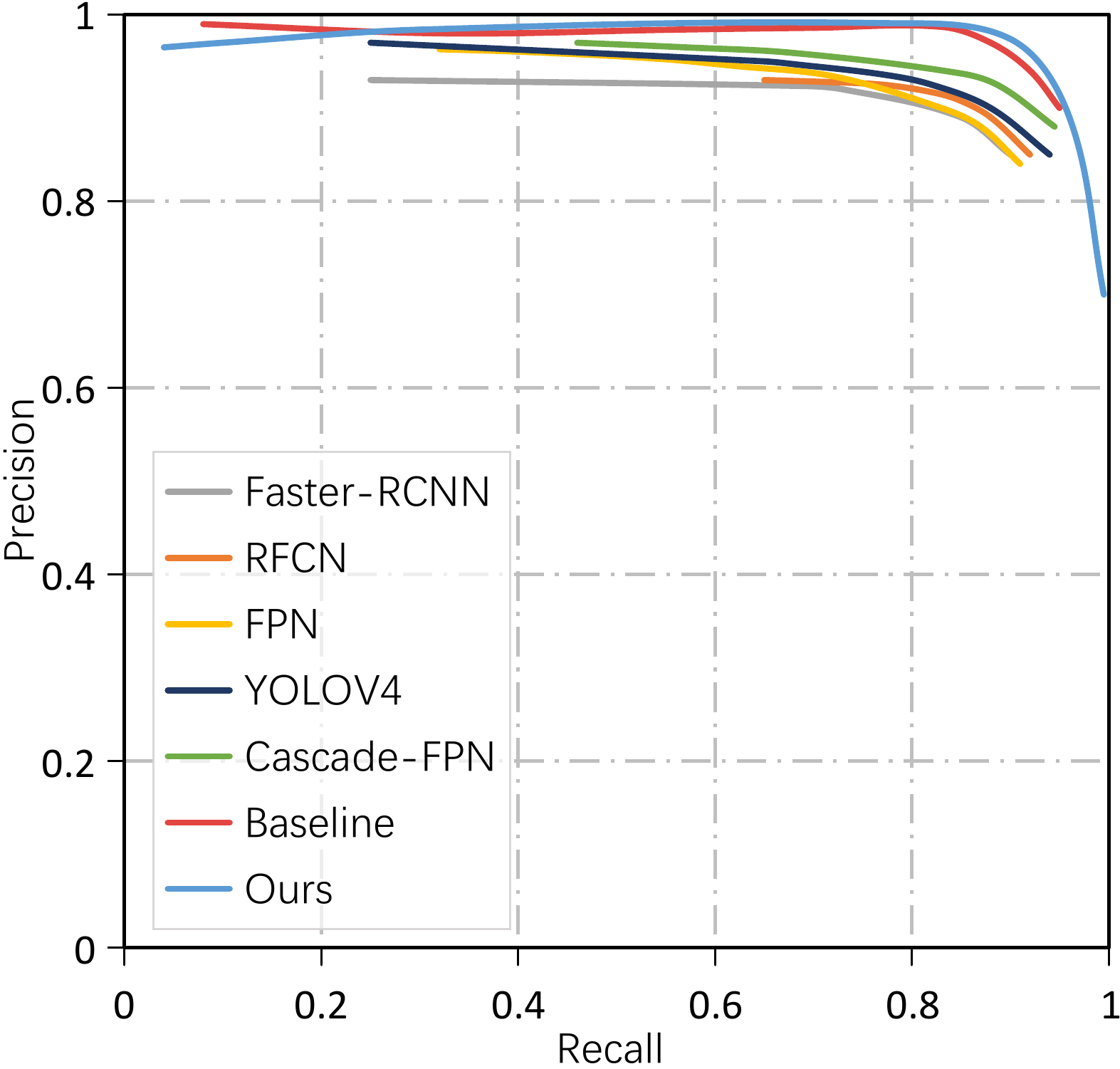}}
	\subfigure[Pothole]{\includegraphics[scale=0.27]{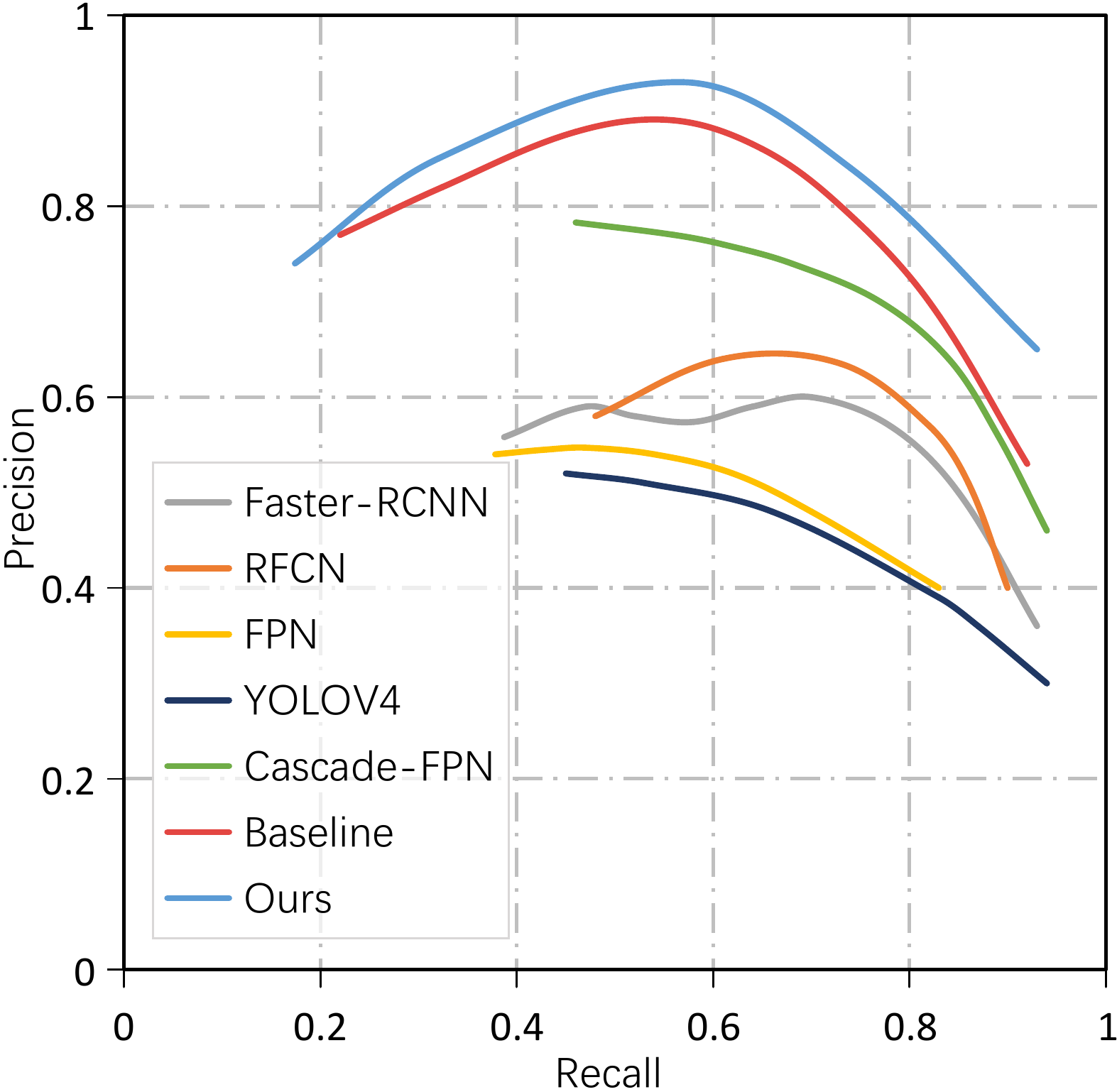}}
	\caption{The Precision-Recall curves of compared methods on four categories. Six methods are included for comparison, in which the backbone network of Faster-RCNN is ResNet101.}
	\label{fig:6}
\end{figure*}
\begin{figure}[h]
	\centering
	\includegraphics[scale=0.33]{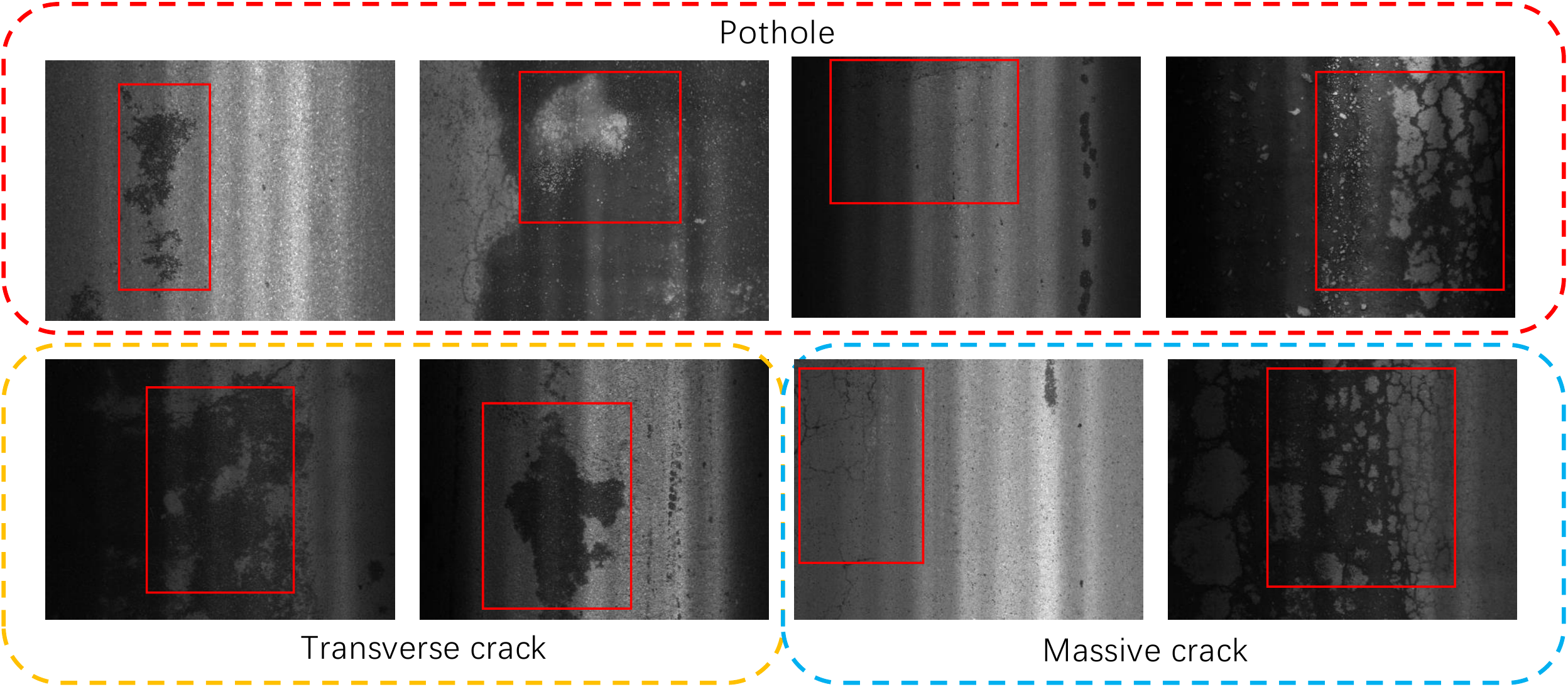}
	\caption{Comparison of Pothole and other categories on CQU-BPDD. The images in each column correspond to different categories, but they have similar crack structures.}
	\label{fig:7}
\end{figure}

\subsubsection{Implementation Details}
We implement our method on PyTorch, which is a well-known deep learning framework in object detection. We adopt YOLOv5 \cite{glenn_jocher_2021_4679653} as the baseline and refer to the official training method of YOLOv5. YOLOv5 has three main parts: backbone, neck, and head as shown in Fig. \ref{fig:2}. The backbone is responsible for extracting features on three scales. The neck collects these features from three different scales through upsampling layers, and inputs them into the head. The head predicts the bounding box around the crack and the class probability associated with each bounding box. The number of training epochs is set to be 300 in the training stage. YOLOv5 will automatically adjust the image size to multiples of 32$\times$32, and the adjusted size is 800$\times$704. During training, Adam optimizer and SGD optimizer are used to update the weights of the network. After getting $D_i$, we train the network in the same hyperparameter settings as before.

To make better use of the knowledge of $D_s$, we select five categories in $D_t$ that are similar to $D_s$ for experiments, as shown in Table \ref{tab:2}. We randomly select 200 images for each category in $D_t$ and label them with ImageLabel\footnote {The ImageLabel website: https://github.com/lanbing510/ImageLabel}, which is an open-source annotation tool. Then, after APAGE has been applied in $D_t$, $D_t$ are augmented in six approaches: regular image sharpening, image channel numerical scaling, additive Gaussian noise, rotation, translation, and contrast transformation. $D_s$ is randomly augmented in one of six approaches. Therefore, we obtain 6000 annotated crack images of $D_t$ and 42082 crack images of $D_s$ as the training set. In the test set, 3622 crack images from $D_t$ are used to test the performance of the DDACDN model, which are corresponding to the categories in Table \ref{tab:2}.
\begin{figure}[h]
	\centering
	{\includegraphics[scale=0.25]{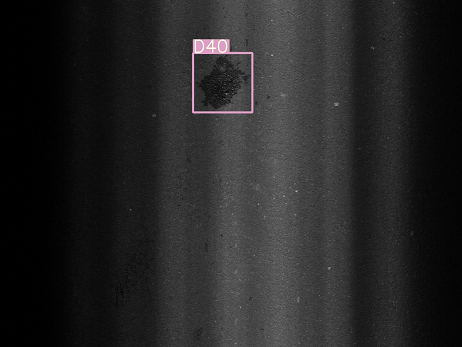}}
	{\includegraphics[scale=0.25]{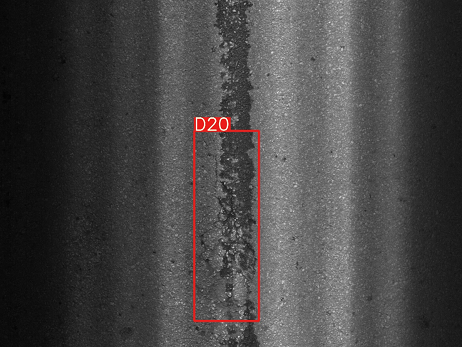}}
	{\includegraphics[scale=0.25]{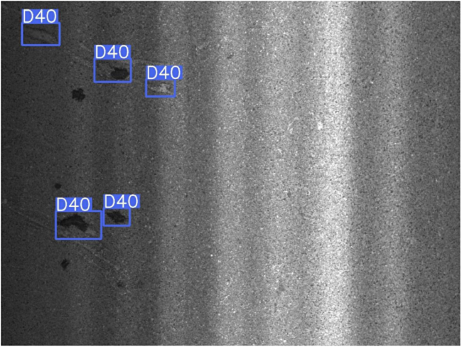}}
	\\
	\vspace{2mm}
	{\includegraphics[scale=0.25]{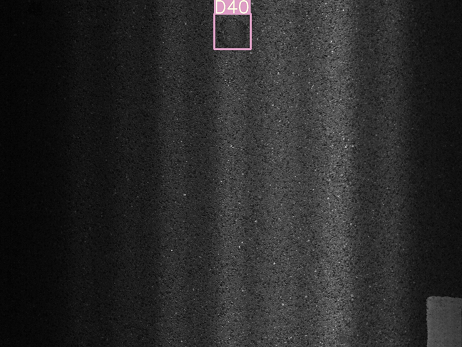}}
	{\includegraphics[scale=0.25]{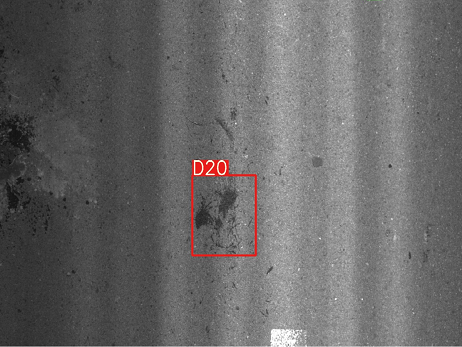}}
	{\includegraphics[scale=0.25]{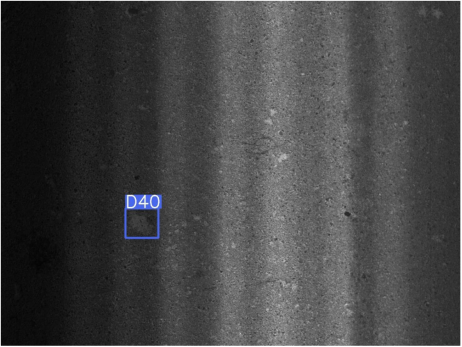}}
	\\
	
	\subfigure[True positives.]{\includegraphics[scale=0.25]{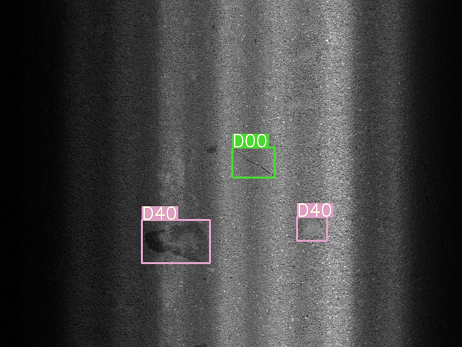}}
	\subfigure[False negatives.]{\includegraphics[scale=0.25]{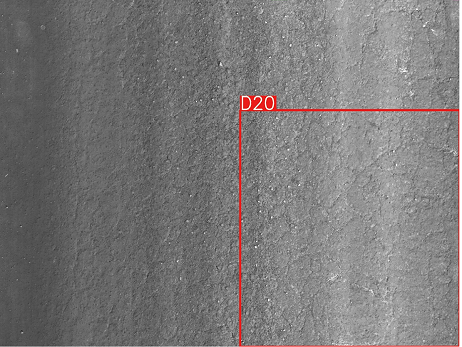}}
	\subfigure[False positives.]{\includegraphics[scale=0.25]{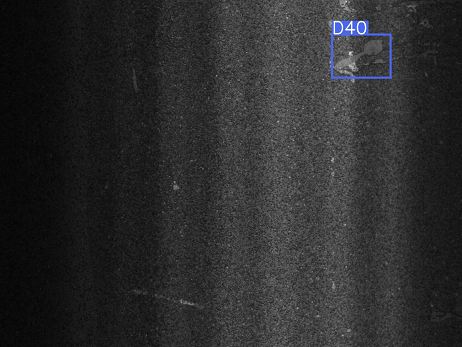}}
	\caption{The true positives, false negatives, and false positives of Pothole. Note that the false negatives are of the Pothole category, which is mistakenly detected as Massive crack. And the false positives are of the Transverse crack category, which is mistakenly detected as Pothole.}
	\label{fig:8}
\end{figure}

\begin{figure*}[h]
	\centering
	\subfigure[Longitudinal crack]{	\begin{minipage}[b]{0.18\textwidth}{\includegraphics[width=1\linewidth]{Fig9_LONG1.pdf}}\vspace{3pt}
			{\includegraphics[width=1\linewidth]{Fig9_LONG2.pdf}}\vspace{3pt}
			{\includegraphics[width=1\linewidth]{Fig9_LONG3.pdf}}\vspace{3pt}
			{\includegraphics[width=1\linewidth]{Fig9_LONG4.pdf}}\vspace{3pt}
			{\includegraphics[width=1\linewidth]{Fig9_LONG5.pdf}}\vspace{3pt}
	\end{minipage}}
	\subfigure[Transverse crack]{	\begin{minipage}[b]{0.18\textwidth}{\includegraphics[width=1\linewidth]{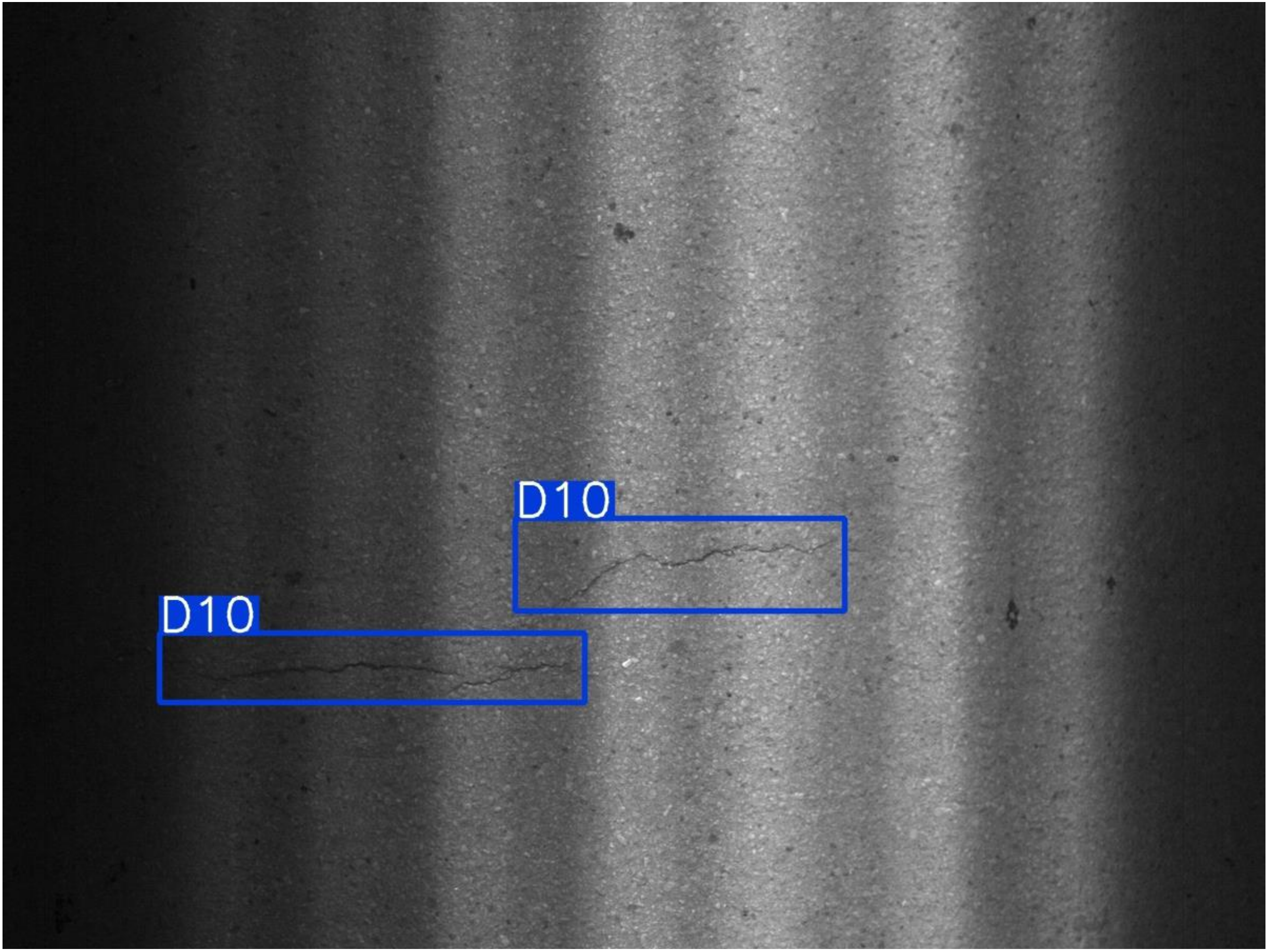}}\vspace{3pt}
			{\includegraphics[width=1\linewidth]{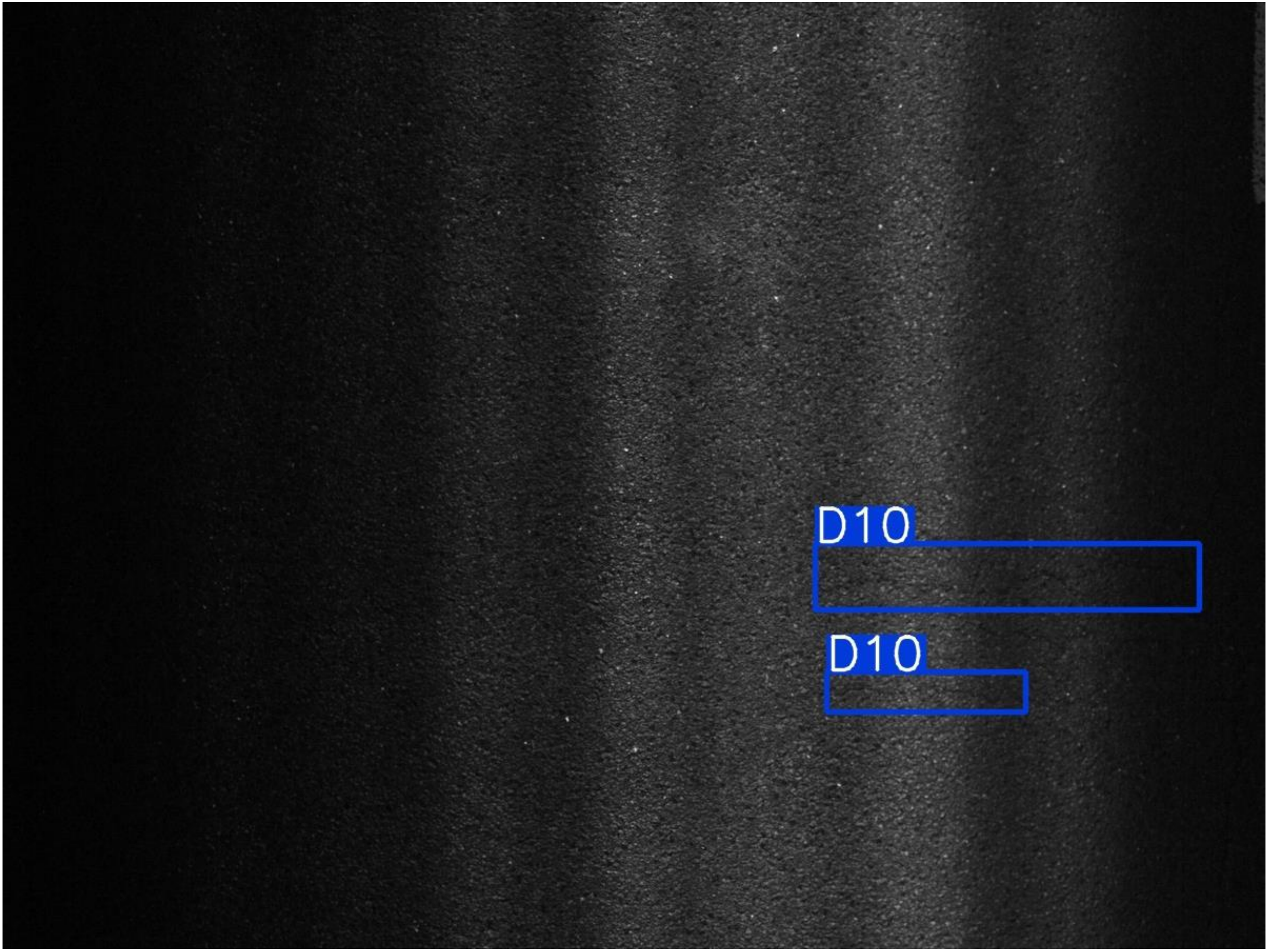}}\vspace{3pt}
			{\includegraphics[width=1\linewidth]{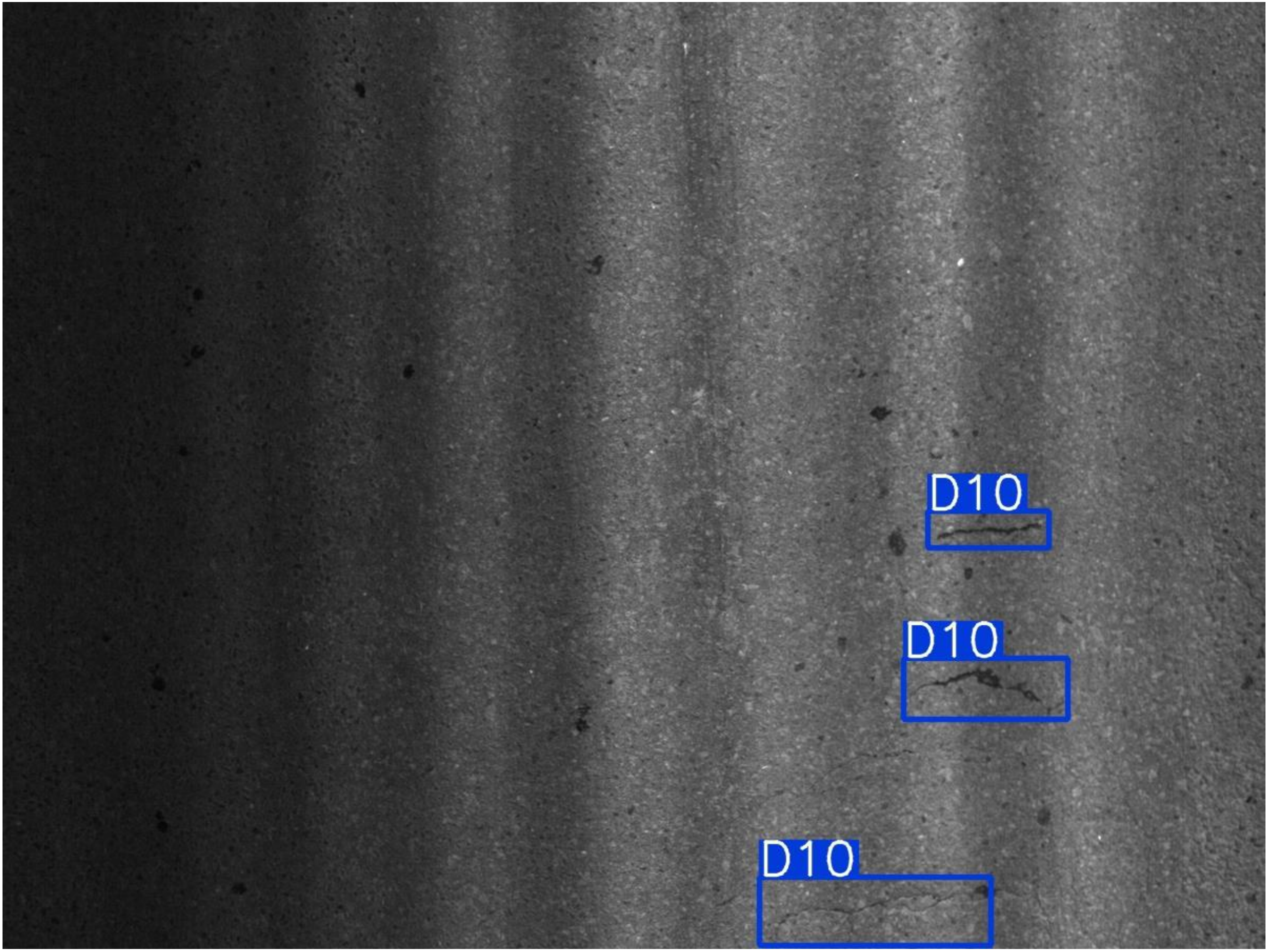}}\vspace{3pt}
			{\includegraphics[width=1\linewidth]{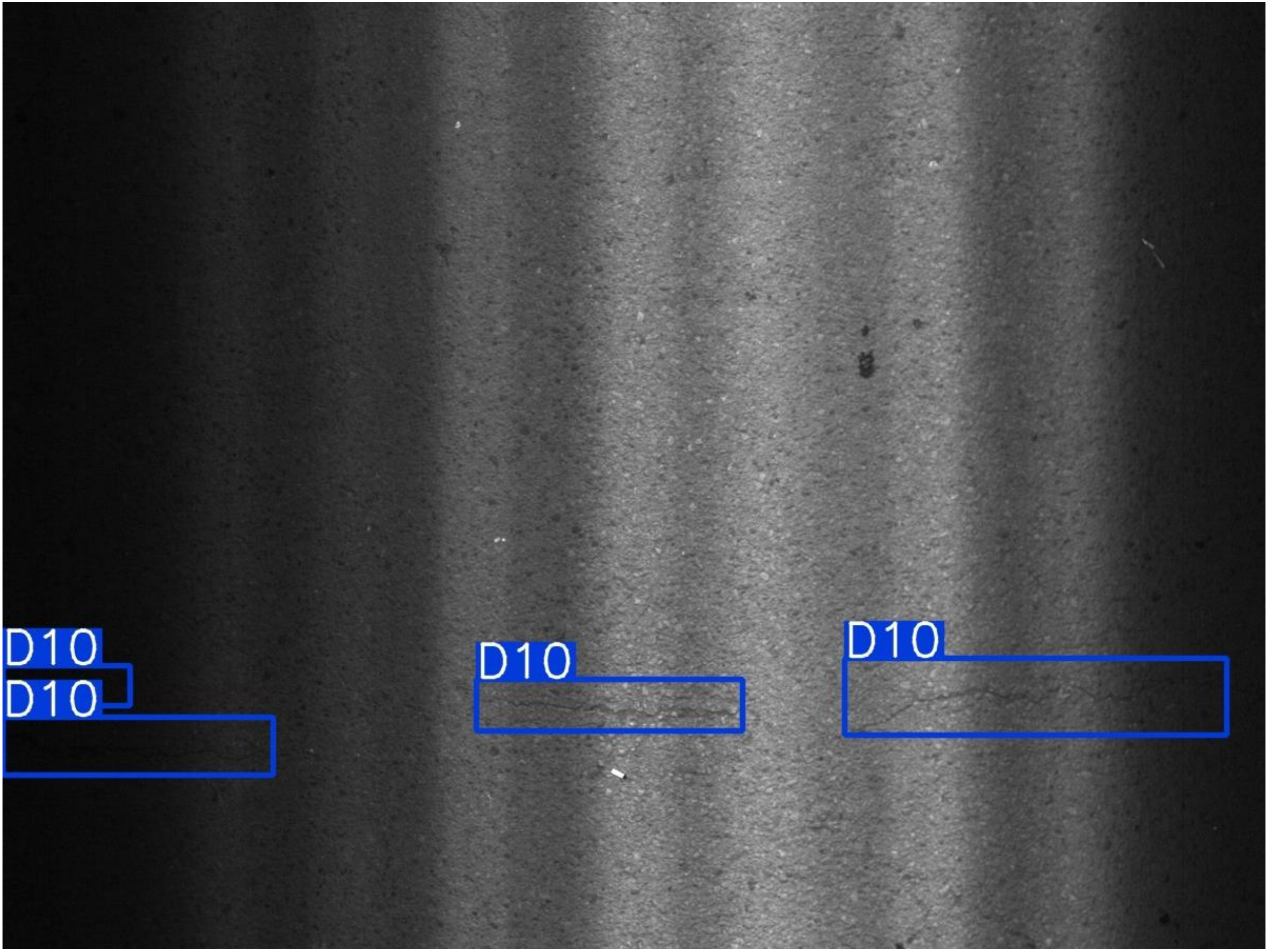}}\vspace{3pt}
			{\includegraphics[width=1\linewidth]{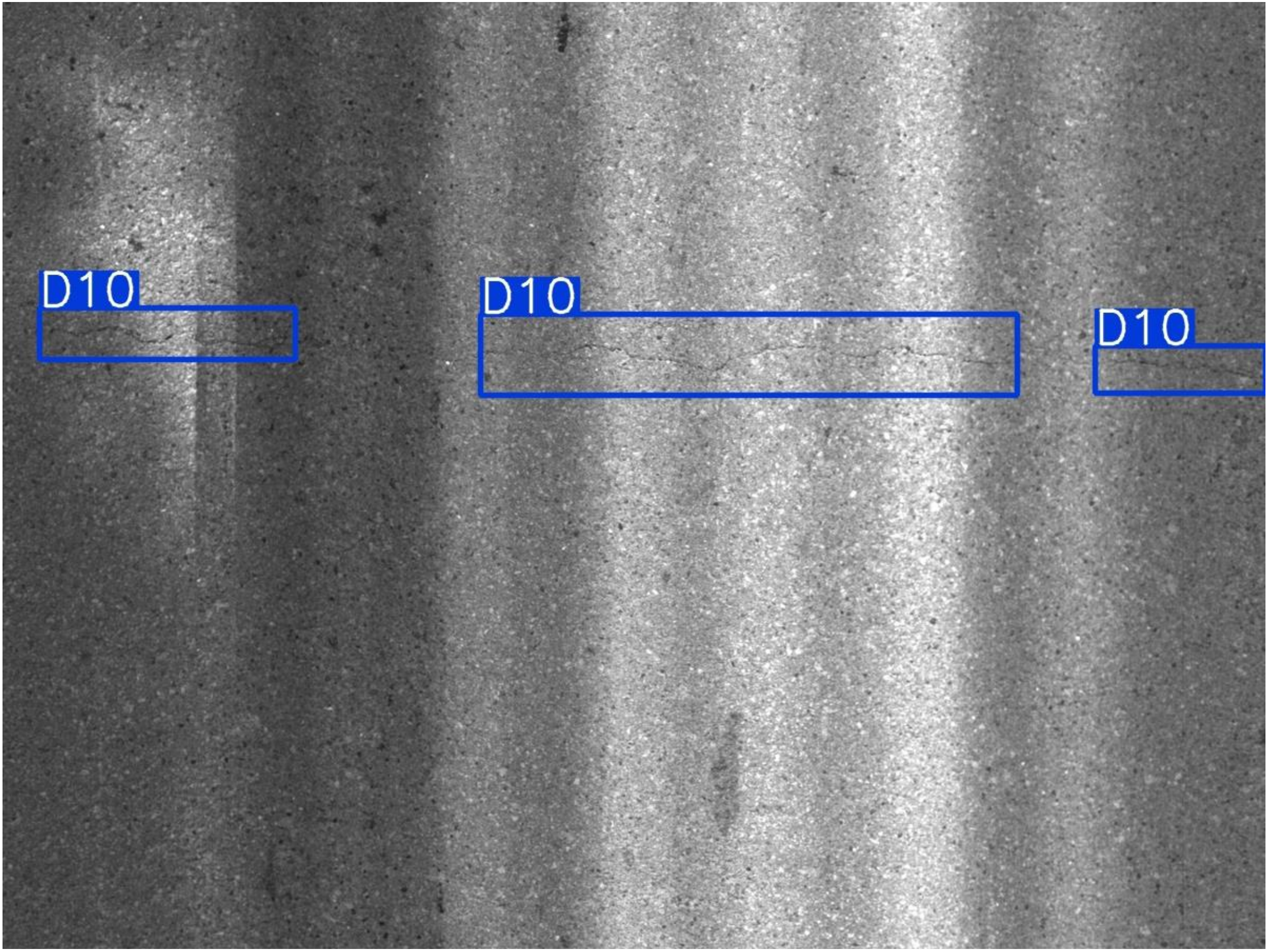}}\vspace{3pt}
	\end{minipage}}
	\subfigure[Massive crack]{	\begin{minipage}[b]{0.18\textwidth}{\includegraphics[width=1\linewidth]{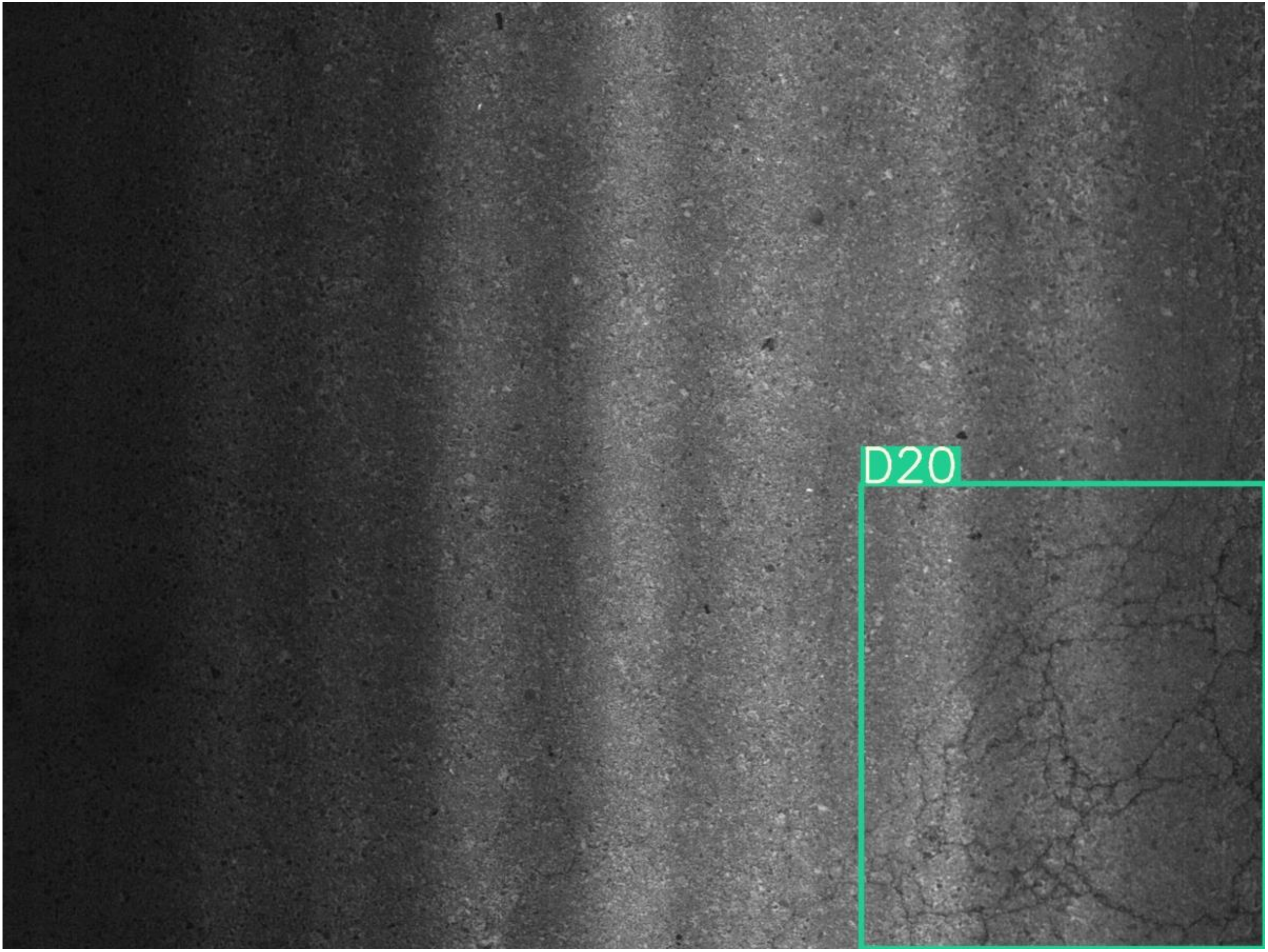}}\vspace{3pt}
			{\includegraphics[width=1\linewidth]{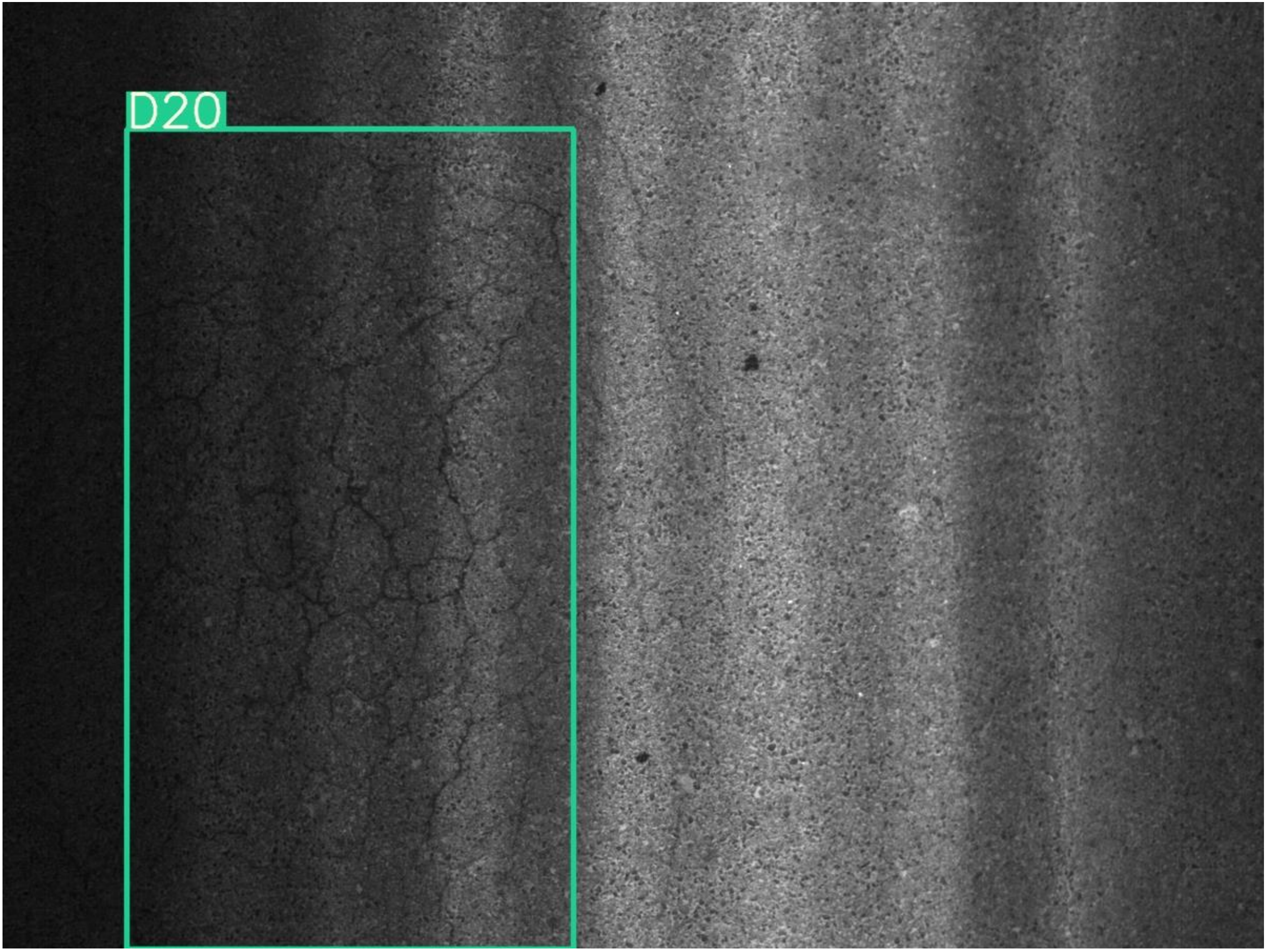}}\vspace{3pt}
			{\includegraphics[width=1\linewidth]{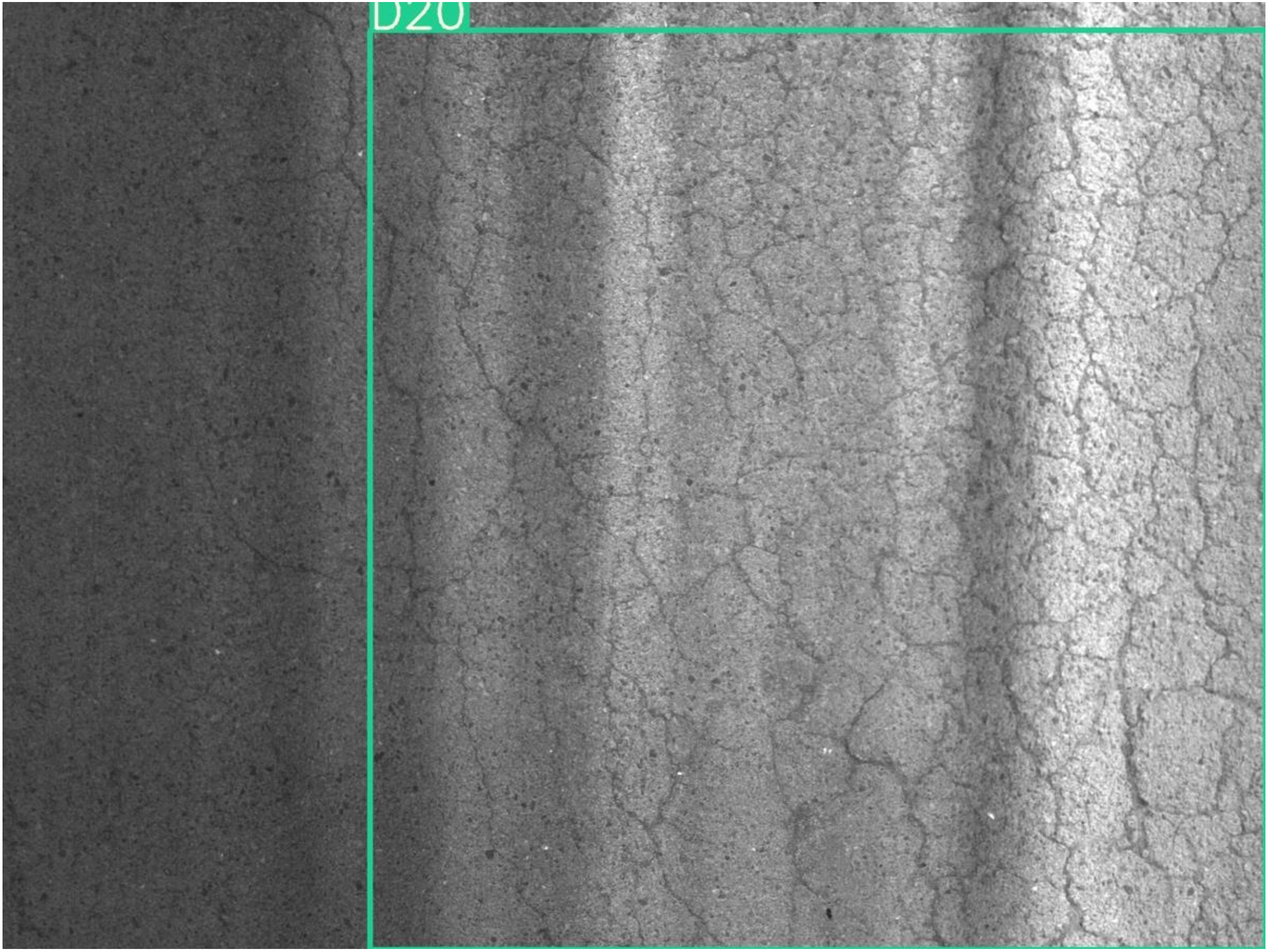}}\vspace{3pt}
			{\includegraphics[width=1\linewidth]{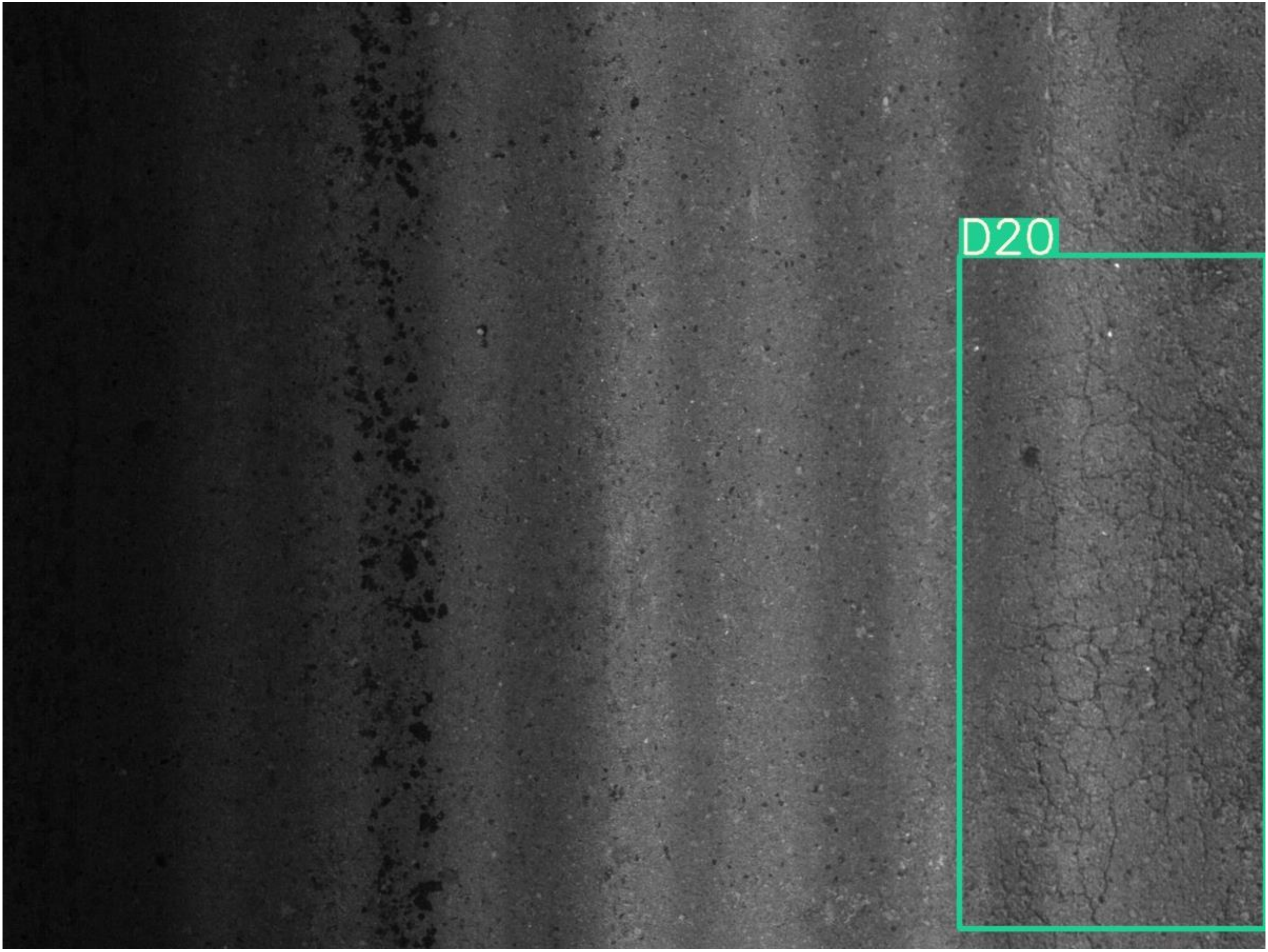}}\vspace{3pt}
			{\includegraphics[width=1\linewidth]{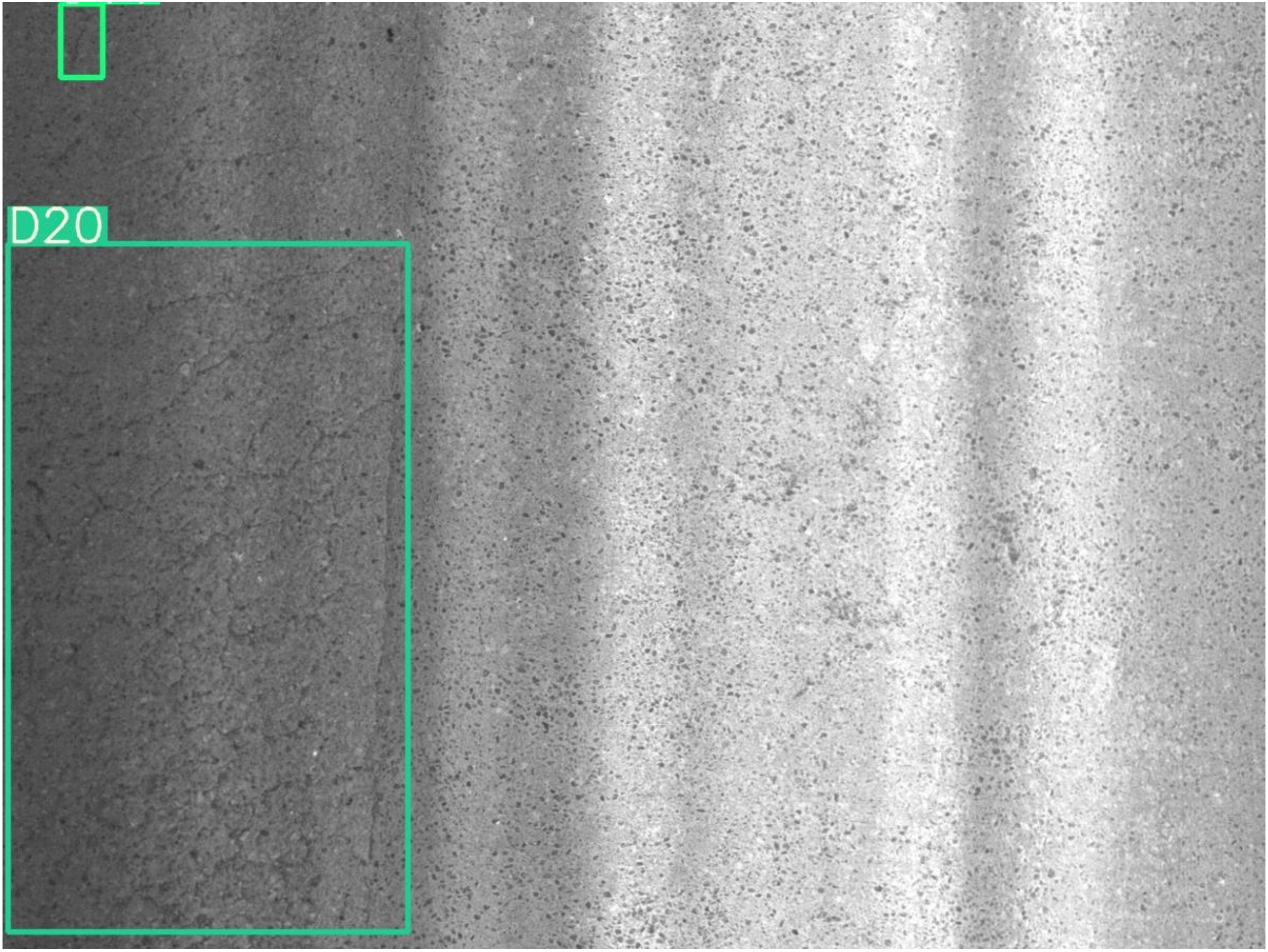}}\vspace{3pt}
	\end{minipage}}
	\subfigure[Crack]{	\begin{minipage}[b]{0.18\textwidth}{\includegraphics[width=1\linewidth]{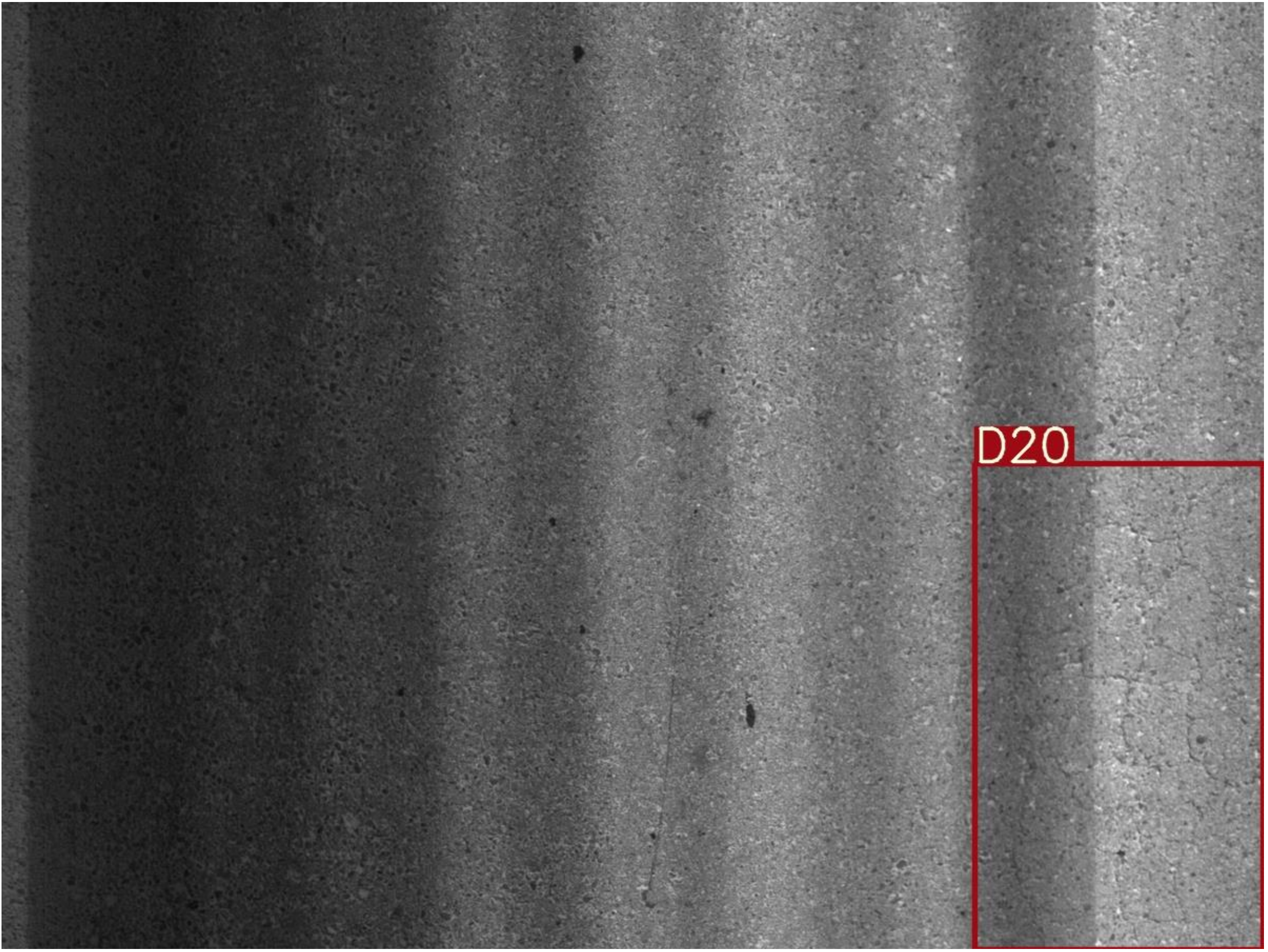}}\vspace{3pt}
			{\includegraphics[width=1\linewidth]{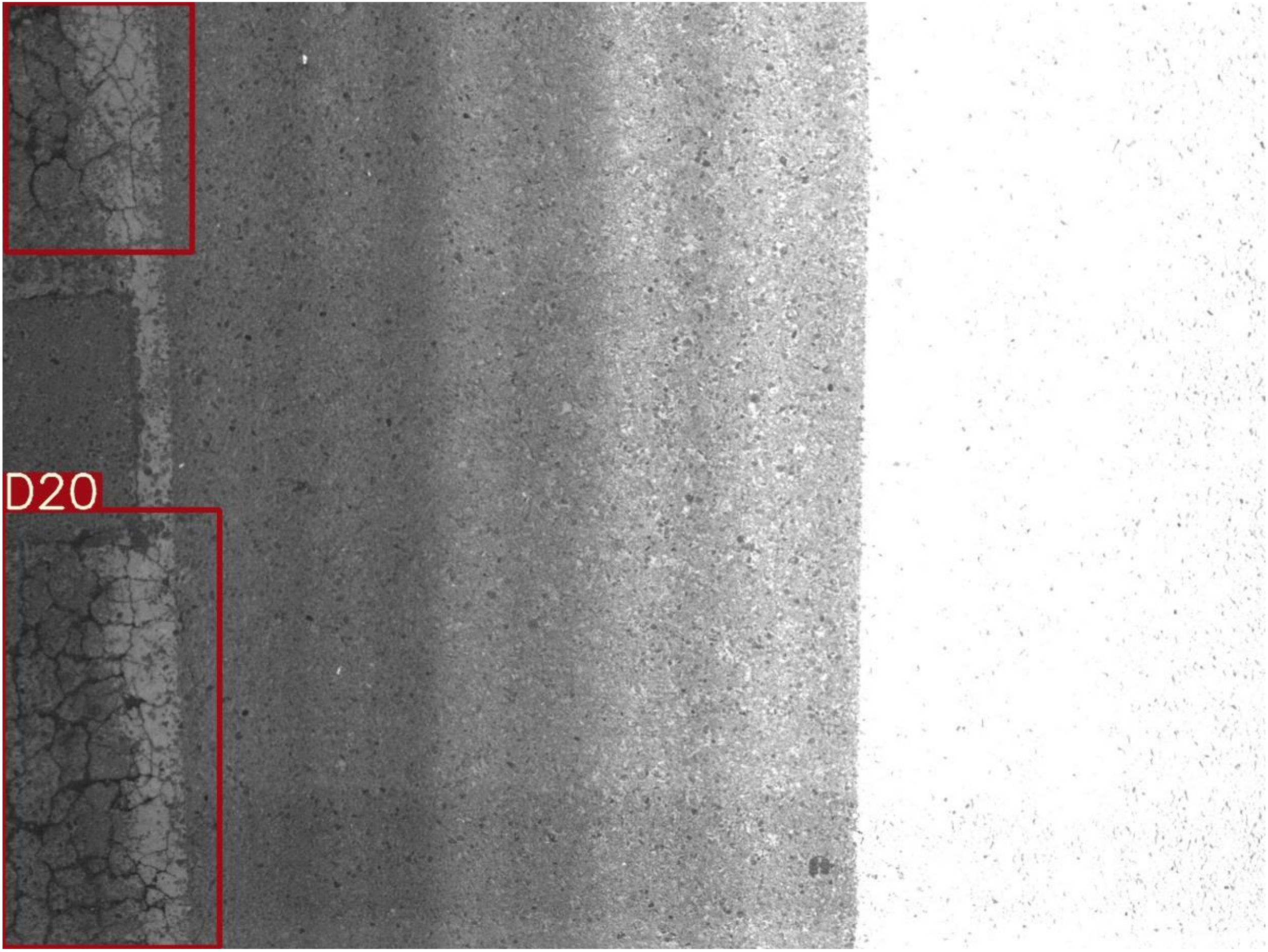}}\vspace{3pt}
			{\includegraphics[width=1\linewidth]{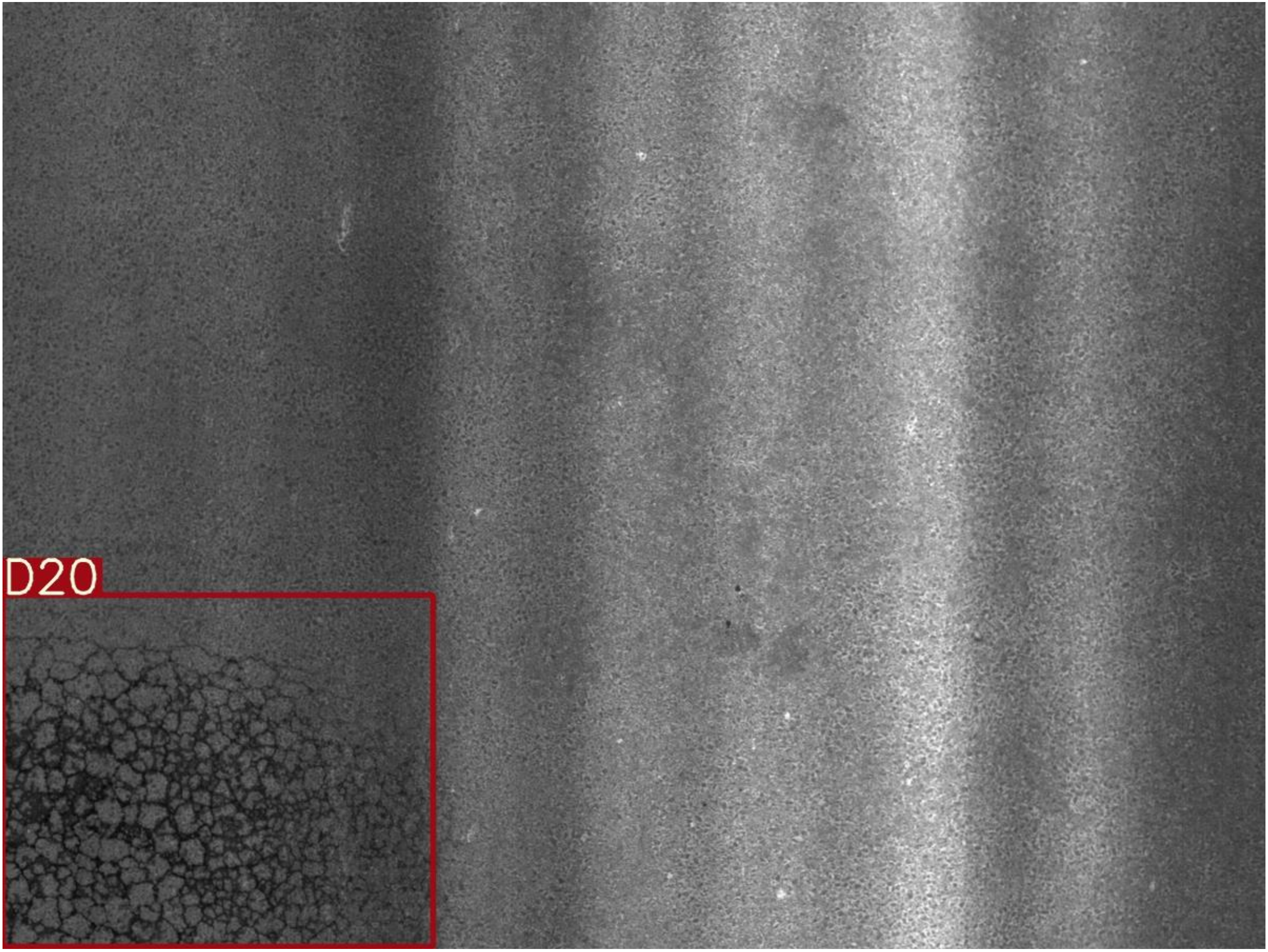}}\vspace{3pt}
			{\includegraphics[width=1\linewidth]{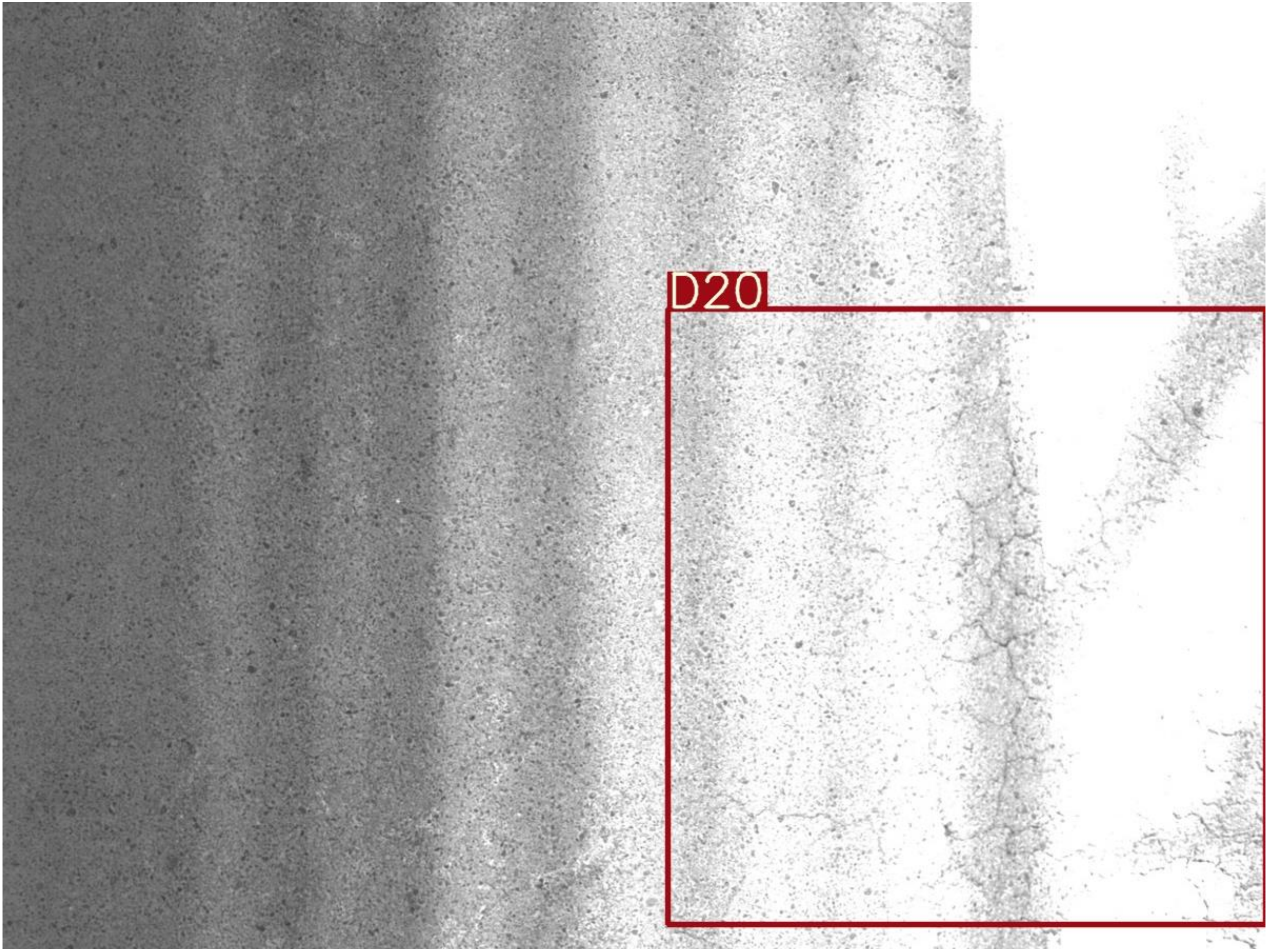}}\vspace{3pt}
			{\includegraphics[width=1\linewidth]{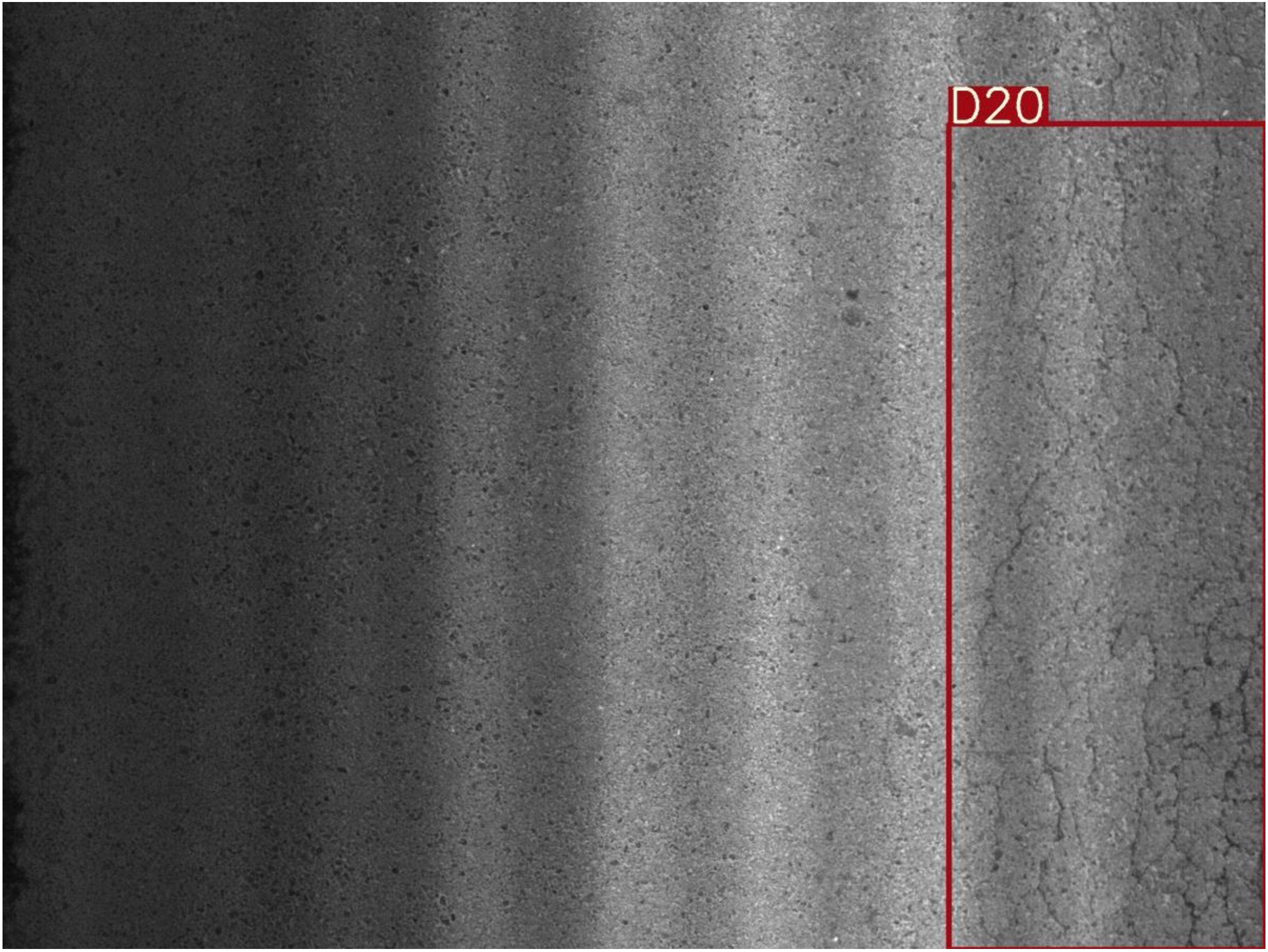}}\vspace{3pt}
	\end{minipage}}
	\subfigure[Loose]{	\begin{minipage}[b]{0.18\textwidth}{\includegraphics[width=1\linewidth]{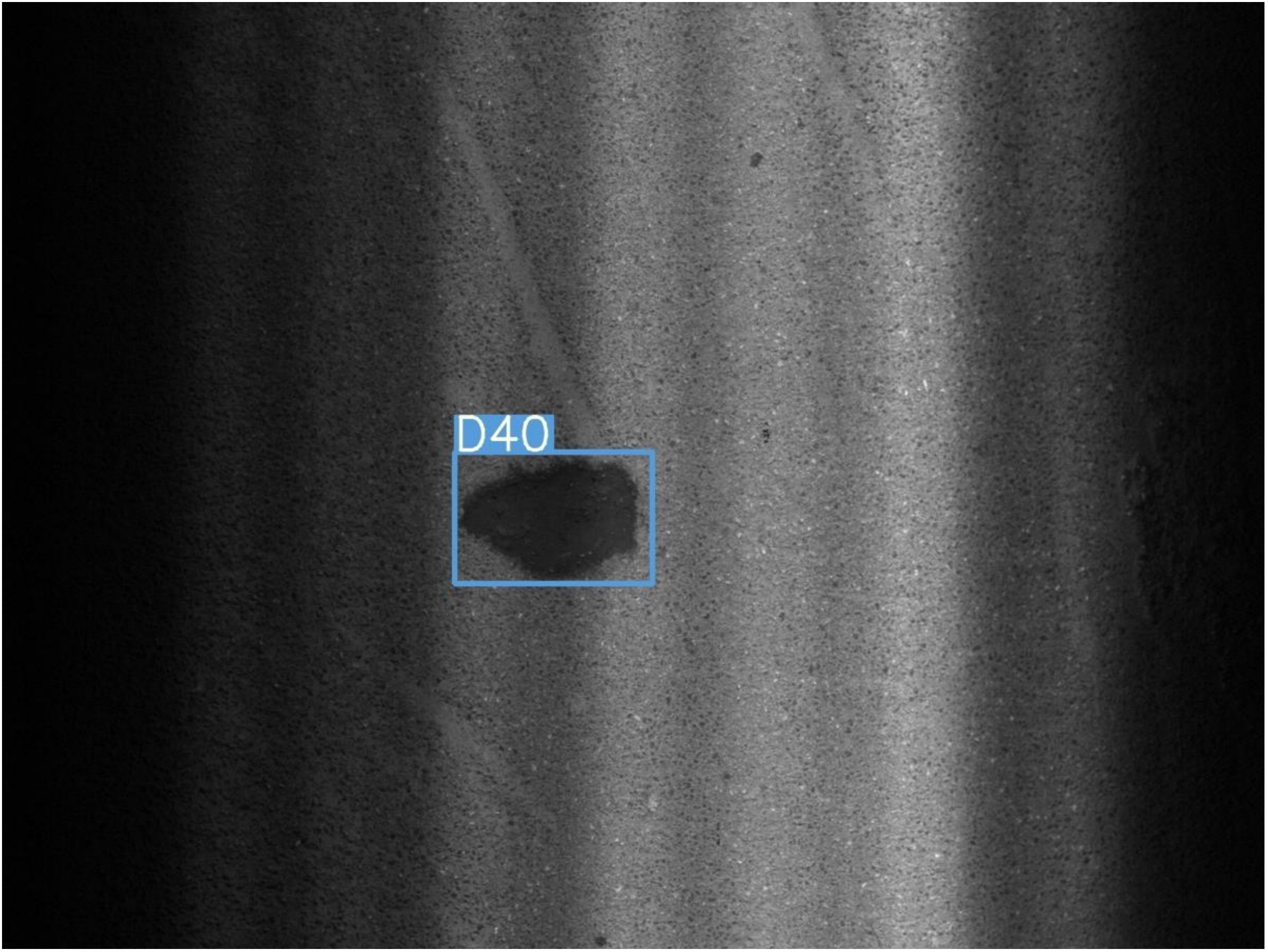}}\vspace{3pt}
			{\includegraphics[width=1\linewidth]{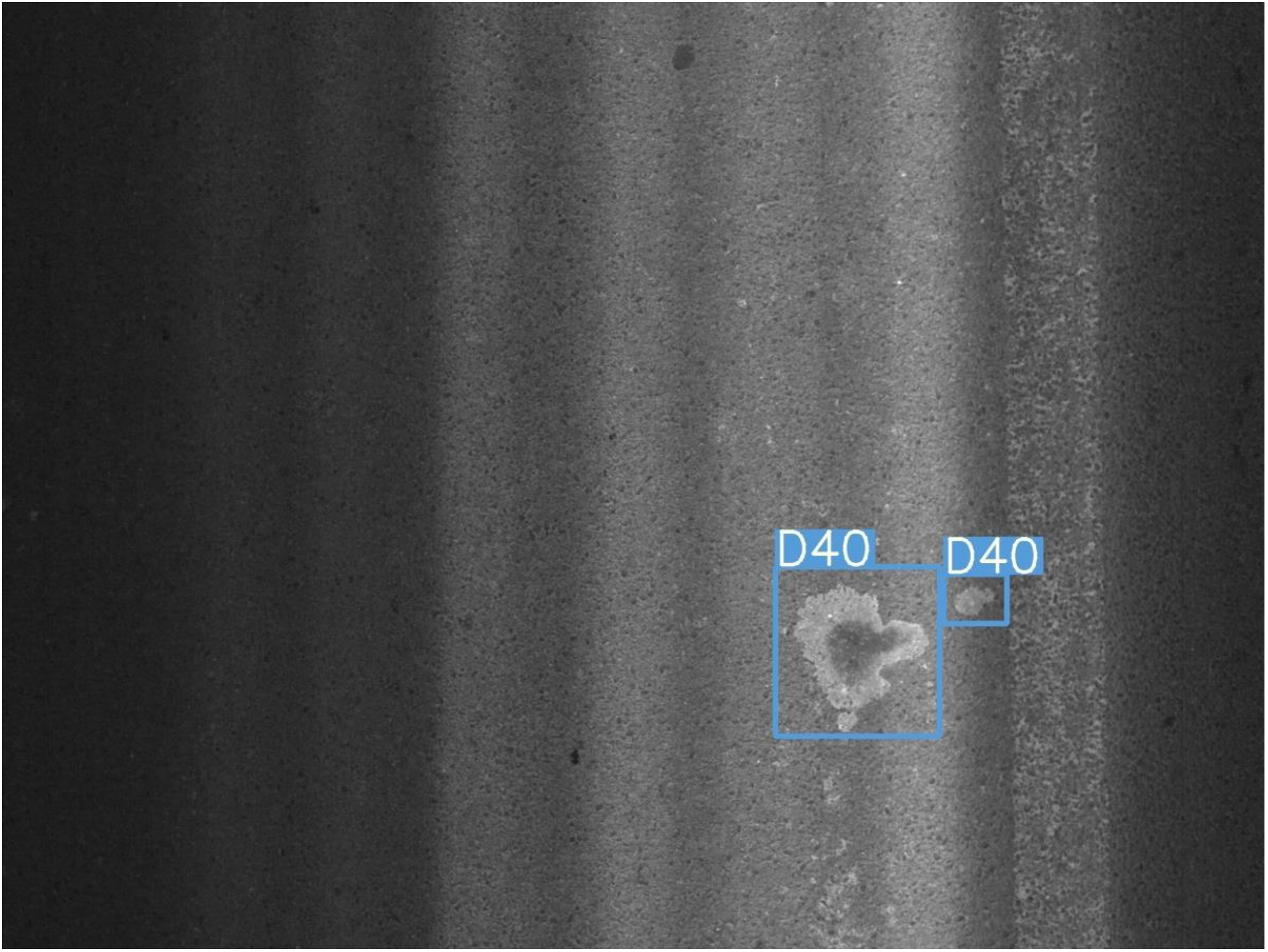}}\vspace{3pt}
			{\includegraphics[width=1\linewidth]{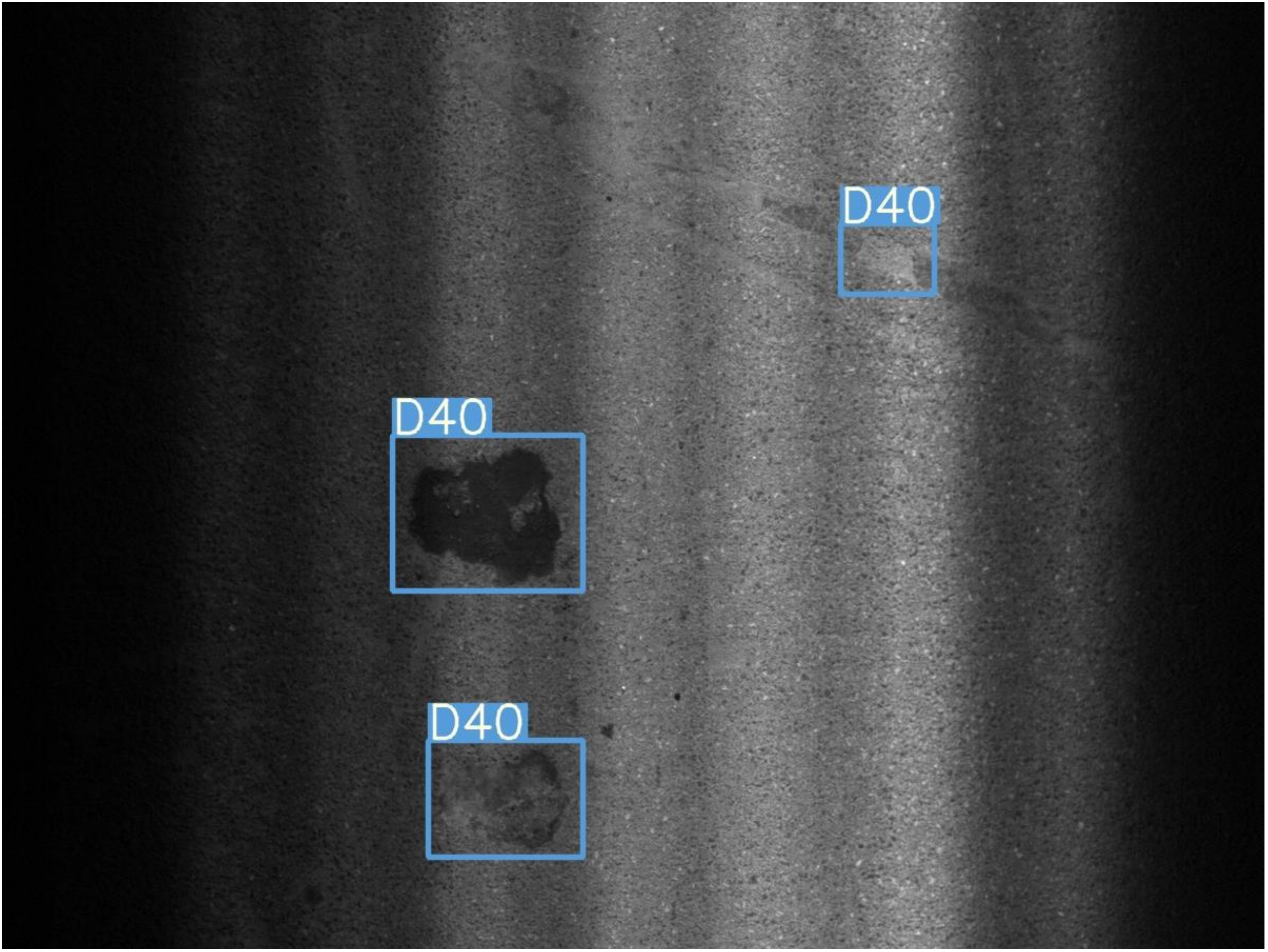}}\vspace{3pt}
			{\includegraphics[width=1\linewidth]{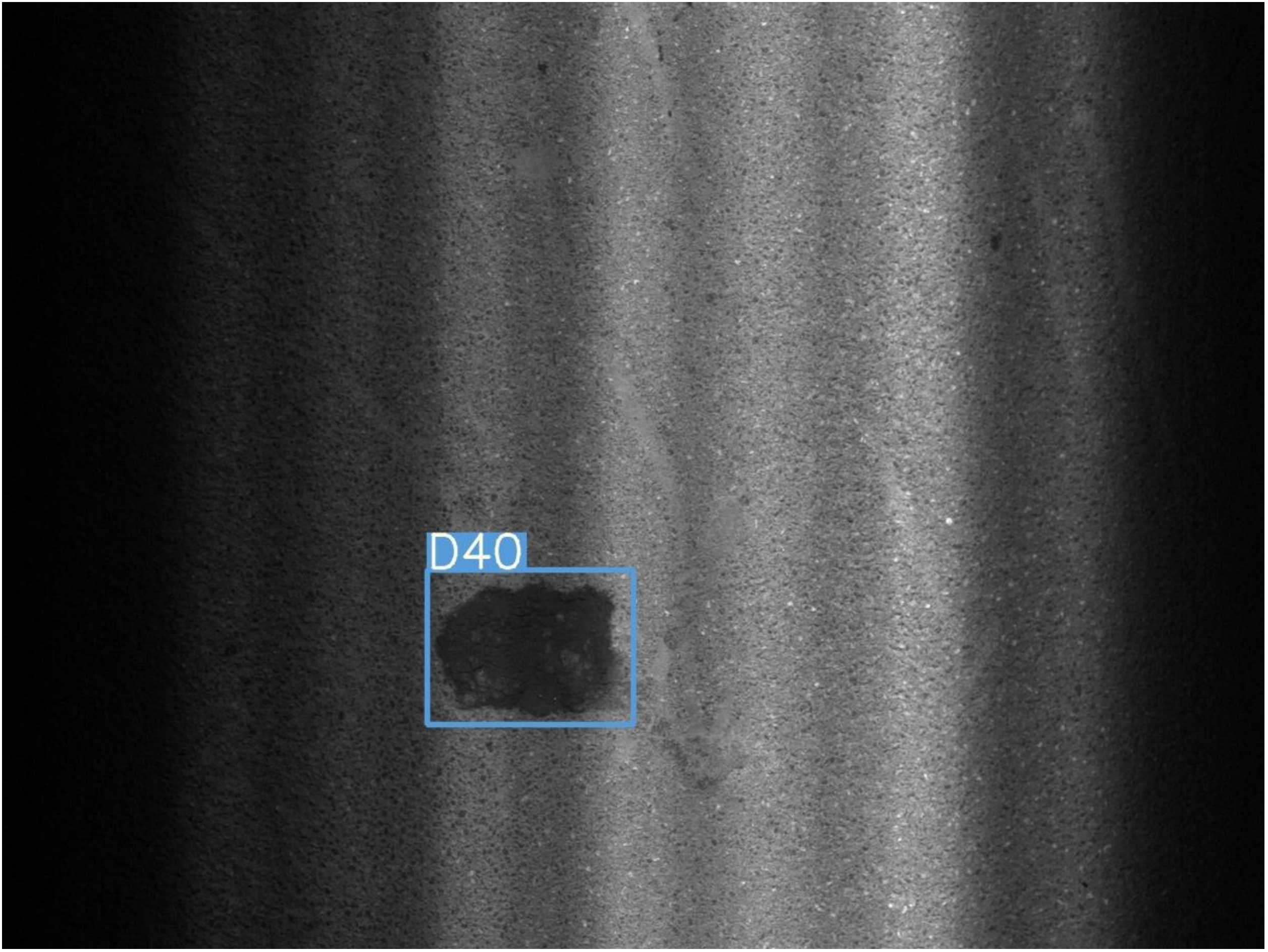}}\vspace{3pt}
			{\includegraphics[width=1\linewidth]{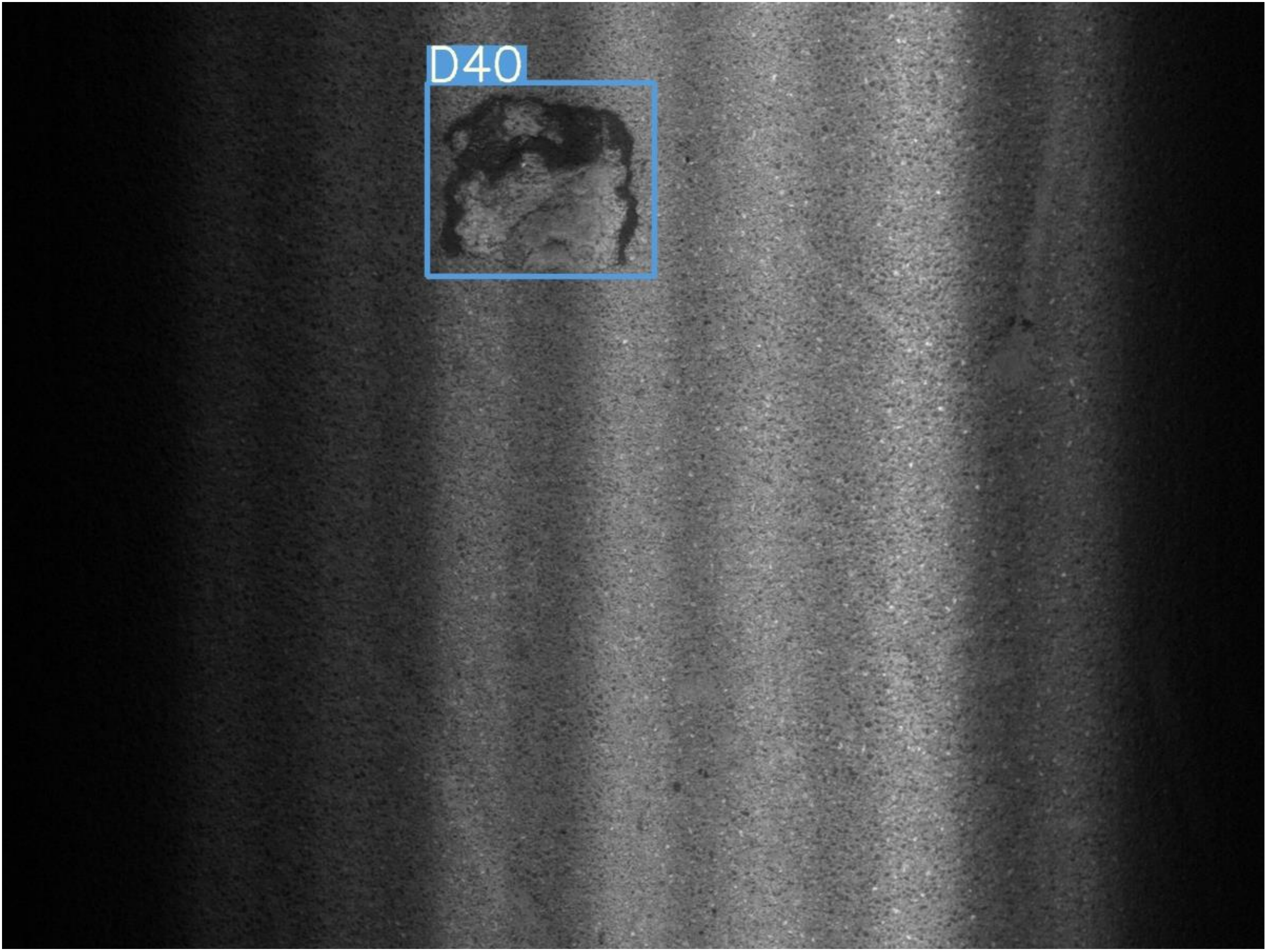}}\vspace{3pt}
	\end{minipage}}
	\caption{Part of the test results on the CQU-BPDD test set. D00, D10, and D40 represent Longitudinal crack, Transverse crack, and Loose respectively. In particular, D20 represents Crack and Massive crack. In order to facilitate discrimination, our detection box uses different colors to distinguish each category.}
	\label{fig:9}
\end{figure*}

\subsubsection{Compared Methods}
We compare the DDACDN model with state-of-the-art methods, which are all trained on the training dataset, and test them on the $D_t$ test set. In general, ten well-known methods are used to detect cracks. The first six are two-stage detection methods, including Faster-RCNN \cite{ren2016faster} with ResNet50, ResNet101 \cite{he2016deep}, and VGG16 backbone networks, RFCN \cite{dai2016r}, FPN \cite{lin2017feature}, and Cascade-FPN \cite{cai2018cascade}. The next two are transfer learning methods, namely DA-Faster-RCNN and DA-Faster-ICR-CCR. The latter two are one-stage detection methods, namely YOLOv4 and YOLOv5 (Baseline). All the hyper-parameters involved in the compared methods have been well adjusted. And the IoU threshold in these methods is first increased from 0.1 to 0.9, each time increasing by 0.1, and the value that makes the experiment results the best is taken as the final value of IoU threshold.

\subsection{Results on CQU-BPDD}
Table \ref{tab:1} lists the detection performance of eight deep learning methods and two transfer learning methods. It can be clearly seen from these results that our work consistently outperforms the compared methods on different evaluation metrics. Fig. \ref{fig:6} shows the P-R curves of deep learning methods, YOLOv5 achieves the best performance among compared deep learning methods and serves as the backbone network of our DDACDN model. Even so, our method outperforms YOLOv5 on \textbf{P}, \textbf{R}, \textbf{F$_1$} in the four categories and \textbf{Acc}. Among them, in \textbf{F$_1$}, they have been increased by 5.4\%, 3.6\%, 2.1\%, and 7.9\%, respectively, and \textbf{Acc} has an increase of about 5.7\%. Furthermore, DA-Faster-ICR-CCR \cite{xu2020exploring} performs the best over the two transfer learning methods, and DDACDN also outperforms it on all metrics, with \textbf{F$_1$} score increasing over 2\% for each category. These results demonstrate that the domain invariant features extracted by the model can help crack detection on the target domain, which verifies the effectiveness of our method. In DDACDN model, the \textbf{R} of the Pothole category is lower than Cascade-FPN and RFCN, and all the evaluation metrics of other categories have been significantly improved.
The main possible reason is that our method starts from a multi-scale perspective, so it can better extract detailed feature patterns of cracks. Therefore, DDACDN could better classify cracks in specific categories, e.g., the fine-grained classification patterns can be learned better. However, through observation, we find that some Pothole images incorrectly labeled or similar to other categories, such as massive crack and transverse crack, are classified to crack by our proposed DDACDN, which may cause the low recall for Potholes. The details are shown in Fig. \ref{fig:7}, where each column of cracks is similar, but they are different categories. 

In most scenarios, DDACDN is able to accurately detect and classify crack categories. However, for some particularly difficult or ambiguously labeled crack categories, detection errors may occur. Fig. \ref{fig:8} shows some true positives, false negatives, and false positives of Pothole, where $FN$ and $FP$ are images of Pothole and Transverse crack categories, respectively. We can see that they are mistakenly detected as Massive crack and Pothole. Compared with Massive crack, it can be clearly seen from Fig. \ref{fig:8} that $FN$ has similar crack structures with them, which may lead to a decrease in \textbf{R} of Pothole. Likewise, $FP$ also has similar structures to Pothole. However, the \textbf{P} and \textbf{F$_1$} of the Pothole category have been greatly improved by DDACDN model with more than 6\% and 5.5\% increasement respectively, which illustrates that DDACDN model reduces the number of $FP$ in the Pothole category. 

The overall experimental results indicate that compared to the cascade-based method and the region-based full convolution structure, domain adaptation has a better effect in solving the problem of insufficient information in $D_t$. The visualization of detection results of the DDACDN model on the test set is shown in the Fig. \ref{fig:9}. It is obvious to observe that our method has excellent performance not only in the detection of large cracks, but also in the detection of small cracks.

\subsection{Cross-Dataset Validation}
\begin{figure}[t]
	\centering
	{\includegraphics[scale=0.38]{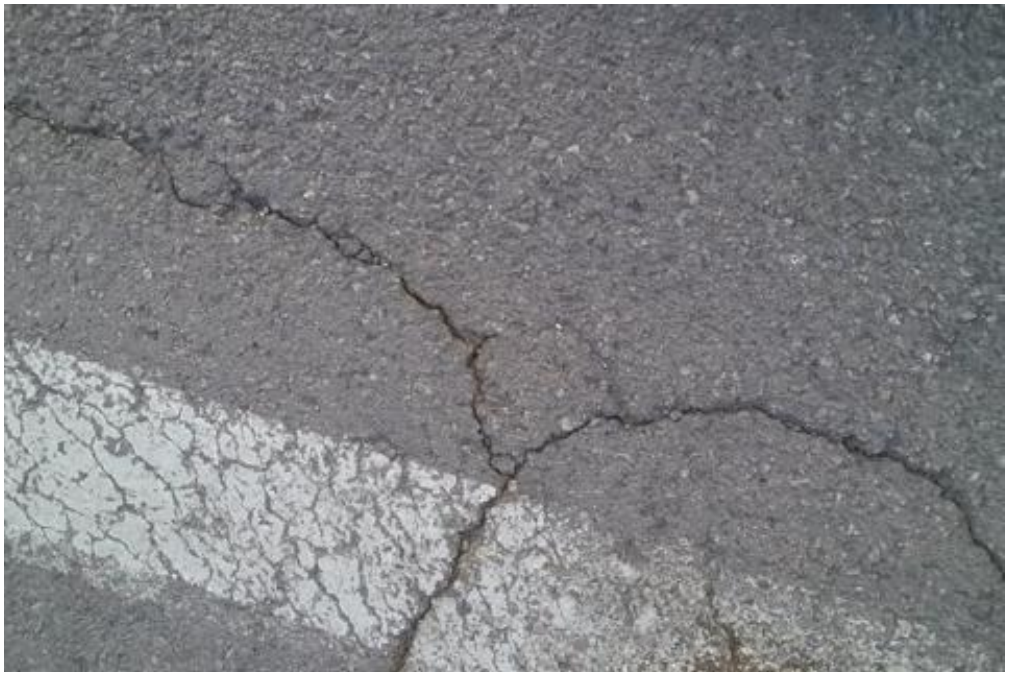}}
	{\includegraphics[scale=0.38]{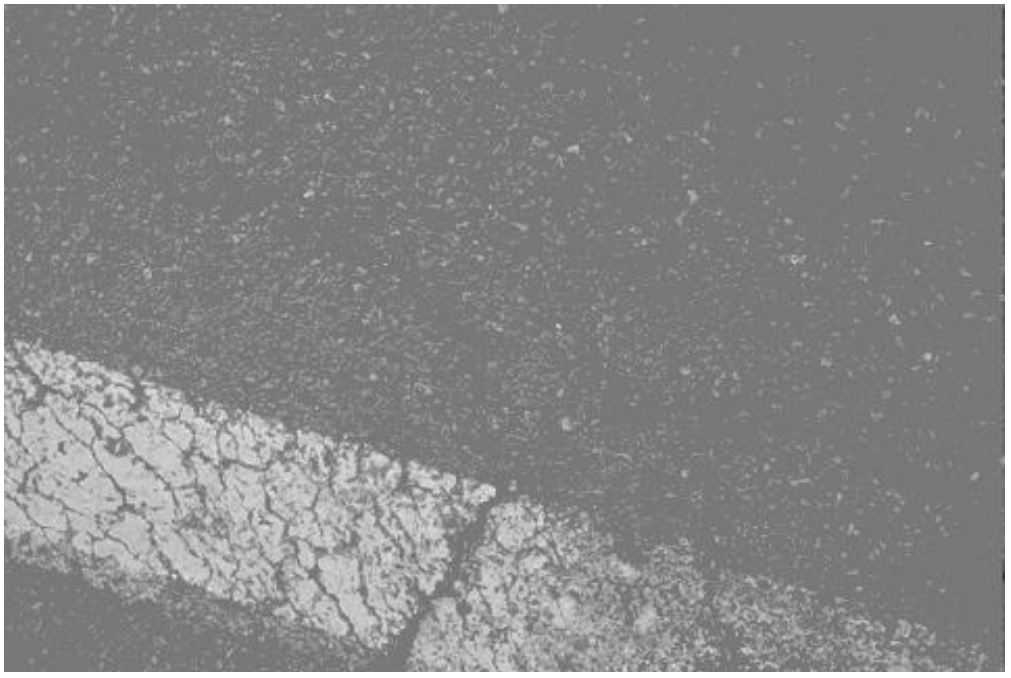}}
	\\
	\subfigure[Diseased images on CFD.]{\includegraphics[scale=0.38]{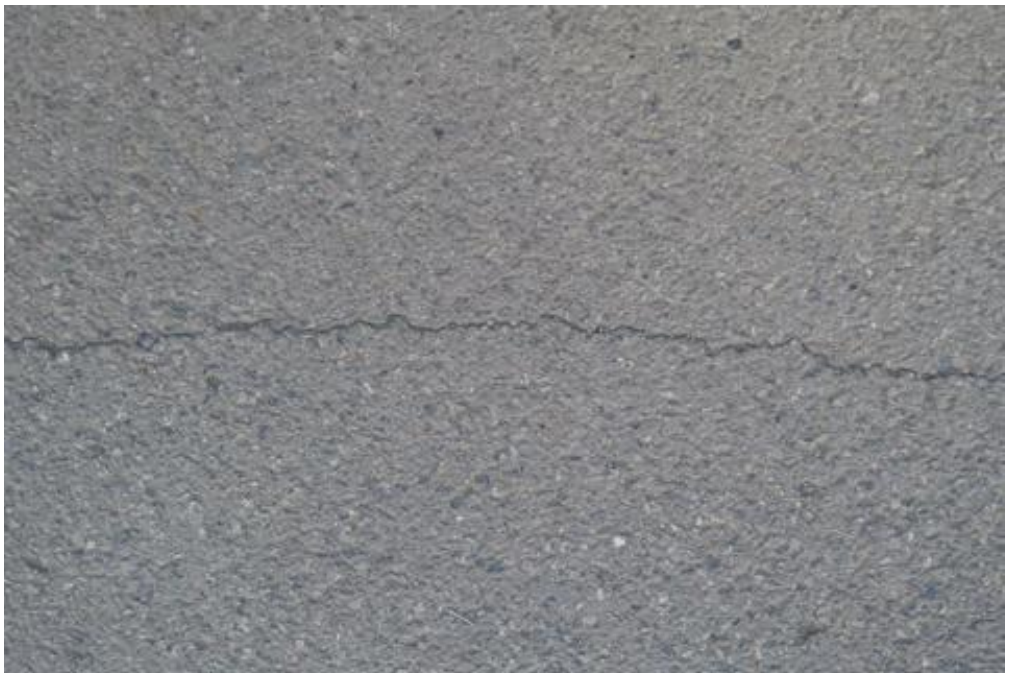}}
	\subfigure[Normal images on CFD.]{\includegraphics[scale=0.38]{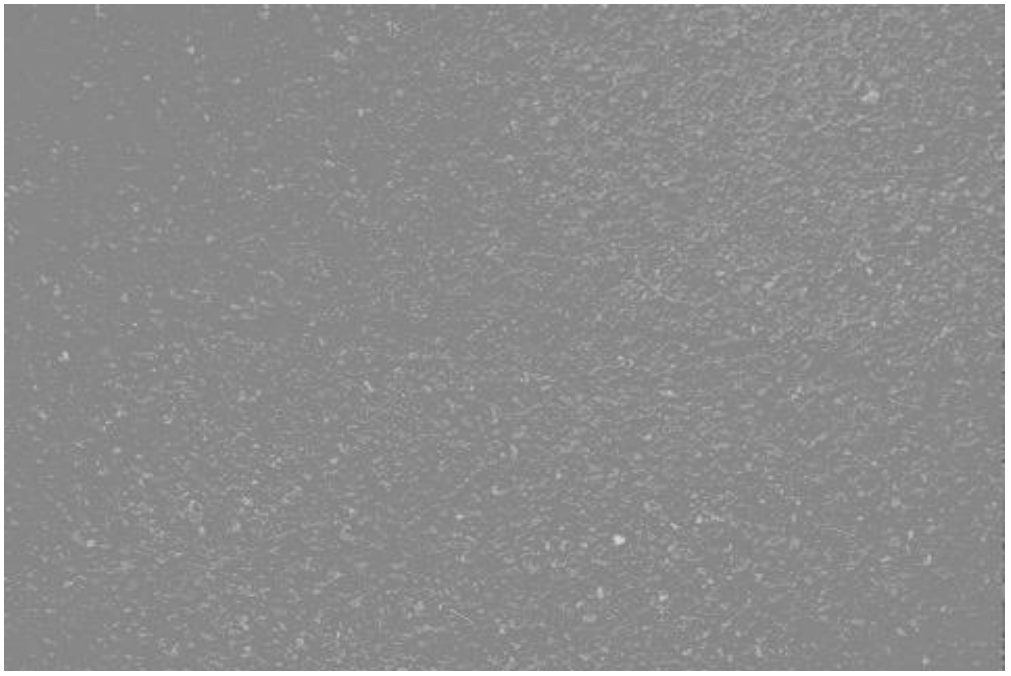}}
	\caption{The pavement image examples on CFD dataset. Note that the normal images are recovered from the disease images via replacing the disease pixels with the average gray value.}
	\label{fig:10}
\end{figure}

In an effort to verify the generalization ability of this method to other data, a commonly used pavement crack segmentation dataset, namely CFD, and CQU-BPMDD are
adopted for validation. However, labels of the diseased categories are different from the category of our experiment. On CFD dataset, we select these 118 diseased images as negative samples and generate positive samples for each image via replacing the disease pixels with their average gray values. Normal images generated in the method are not always applicable. Therefore, we manually filter out some low-quality generated normal images, and only retain high-quality images. Finally, there are a total of 118 diseased images and 104 restored normal images in CFD dataset, and Fig. \ref{fig:10} shows some examples. However, the CFD dataset alone can not sufficiently verify cross-dataset generalization ability due to the small amount of data and the large proportion of cracks in the image. A larger dataset with smaller cracks is urgently needed for more robust generalization.

Among the eight categories of CQU-BPMDD dataset, two categories correspond to the category of CQU-BPDD dataset, namely Longitudinal crack and Transverse crack, with 3886 and 1074 images respectively. Hence, we only detect these two categories on this dataset. Note that the number of Loose and Massive crack categories is too few to be considered. The images on these datasets are only used for testing. The road disease detection model involved in this section is only trained on $D_s$ and $D_t$ without any fine-tuning of these two datasets. This can well compare the cross-dataset generalization abilities between DDACDN model and baseline. Note that we select the first label in the CQU-BPMDD dataset for cross-data generalization experiments.
\begin{table}[h]
	\caption{The performance comparison between our method and the baseline on CQU-BPMDD and CFD datasets.}
	\label{tab:3}
	\renewcommand\arraystretch{1.5}
	\centering
	\setlength{\tabcolsep}{3mm}{
		\begin{tabular}{ c|  c c c c  }\Xhline{1.2pt}  
			Validation Dataset & Methods &F$_1^C$&F$_1^N$&Acc \\ \hline
			\multirow{2}{*}{CFD \cite{shi2016automatic}}&Baseline&96.2\%&95.7\%&95.9\%\\
			&Ours&\textbf{97.1\%}&\textbf{96.5\%}&\textbf{96.8\%}\\
			\Xhline{1.2pt}
			Validation Dataset & Methods &F$_1^L$&F$_1^T$&Acc \\ \hline
			\multirow{2}{*}{CQU-BPMDD}&Baseline&83.0\%&98.8\%&76.7\%\\
			&Ours&\textbf{87.2\%}&\textbf{99.3\%}&\textbf{82.0\%}\\
			\Xhline{1.2pt}
	\end{tabular}}
\end{table}
\begin{figure*}[h]
	\centering
	{\includegraphics[scale=0.28]{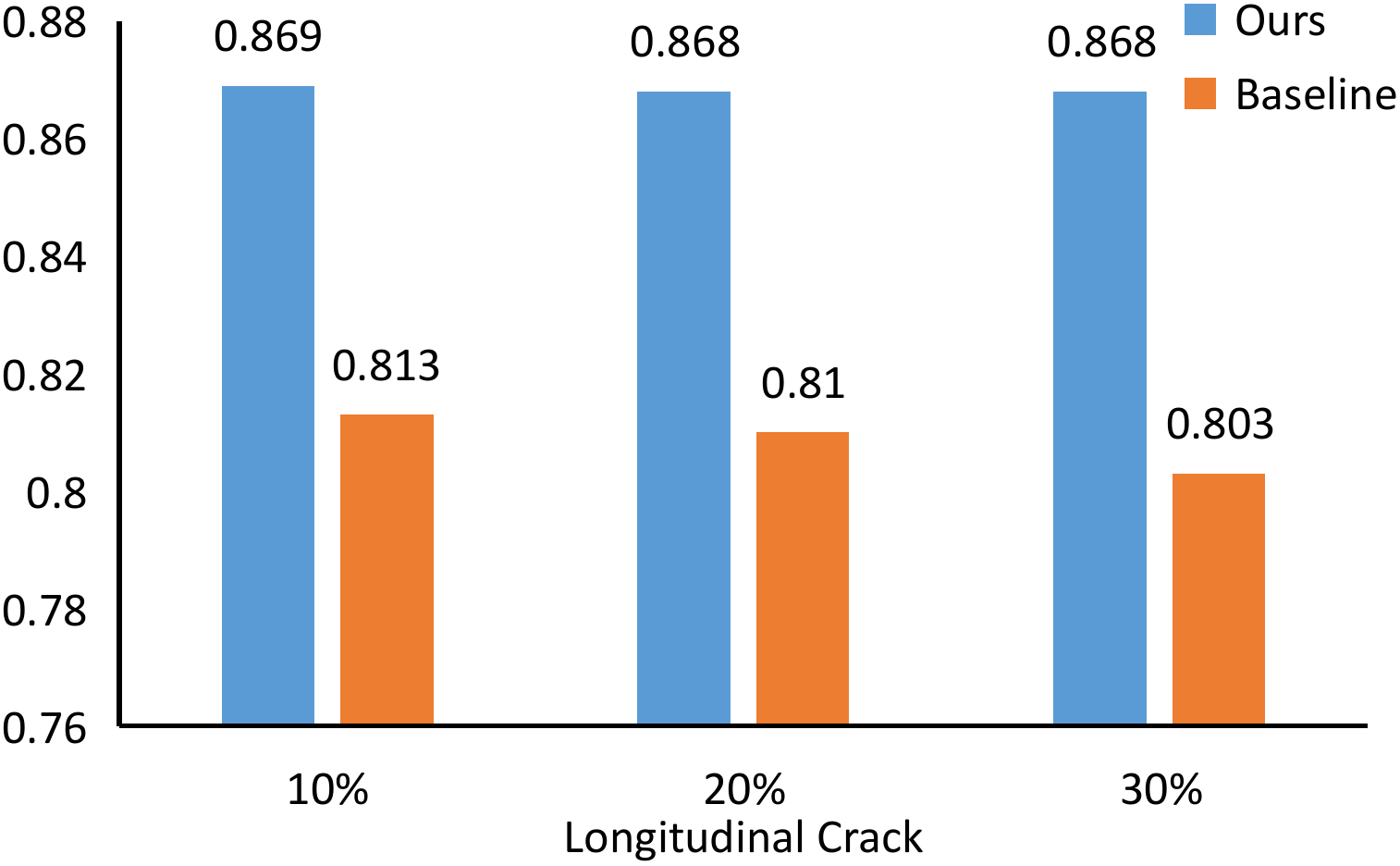}}
	{\includegraphics[scale=0.28]{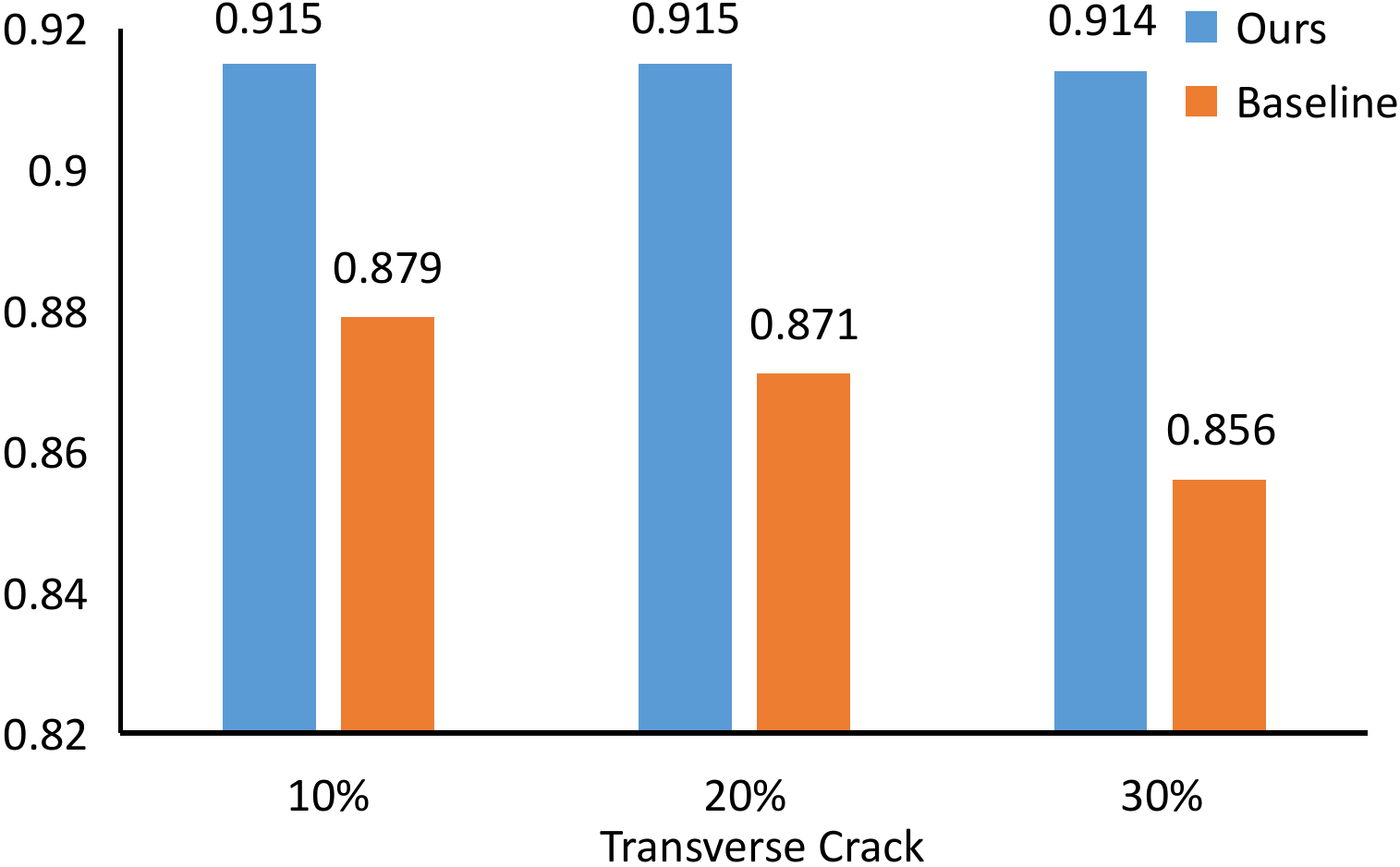}}
	{\includegraphics[scale=0.28]{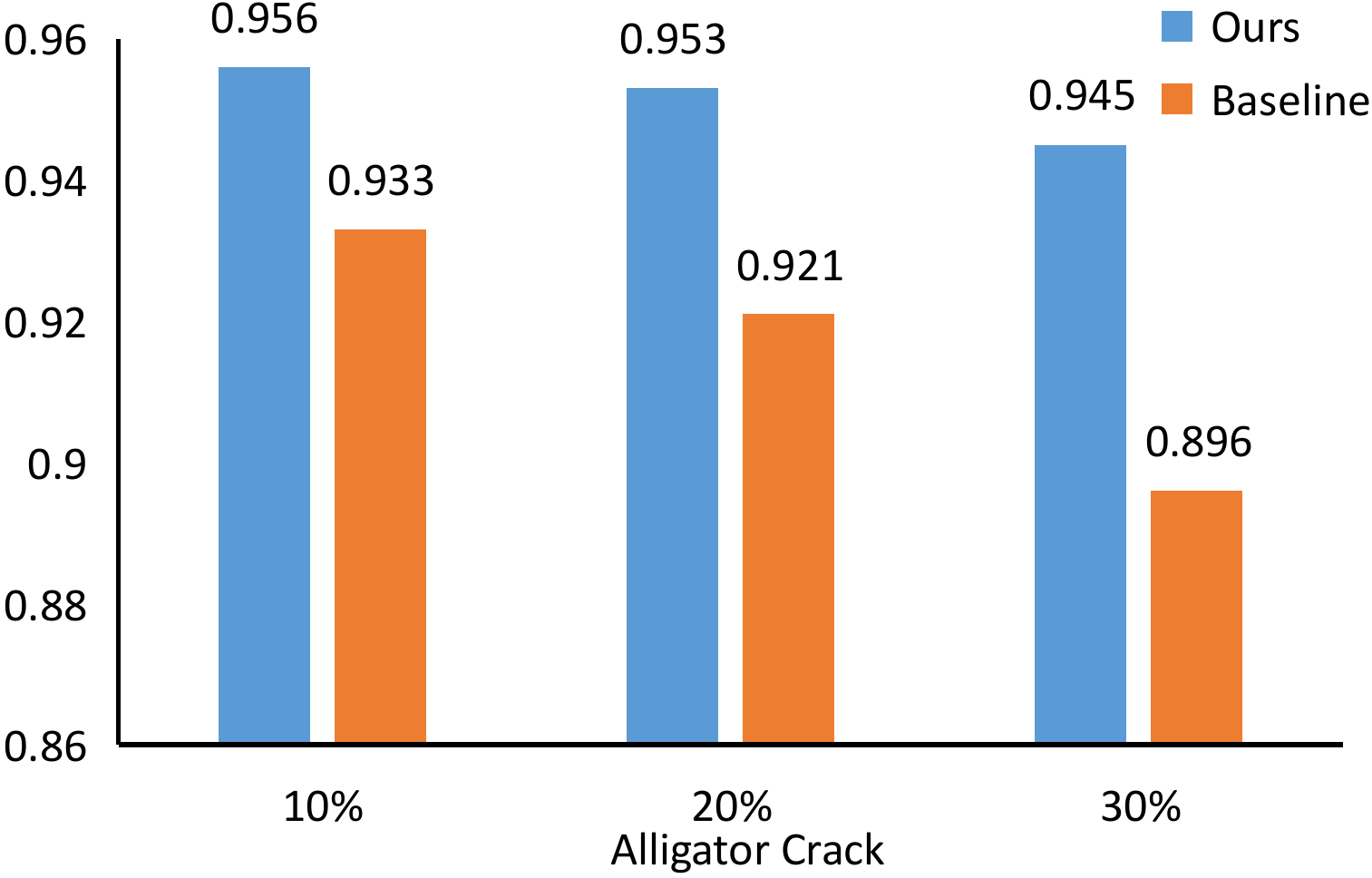}}
	{\includegraphics[scale=0.28]{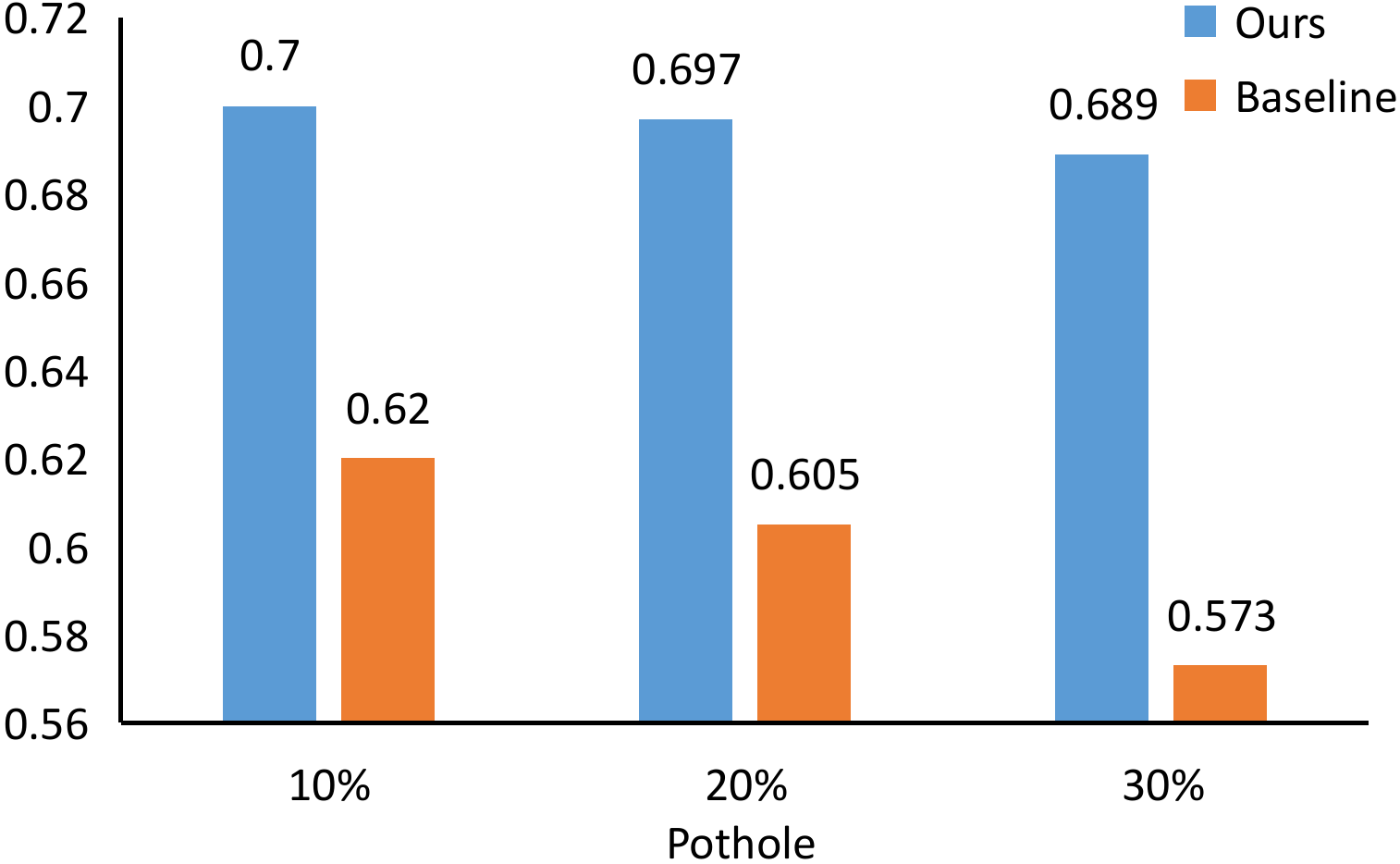}}
	\caption{The performances of DDACDN and Baseline under different noise ratios in F$_1$ of four categories on CQU-BPDD dataset.}
	\label{fig:11}
\end{figure*}

Table \ref{tab:3} shows the performance of DDACDN and Baseline on CFD and CQU-BPMDD datasets, where $F_1^C$, $F_1^N$, $F_1^L$, and $F_1^T$ respectively represent the \textbf{F$_1$} scores of crack, normal, Longitudinal crack and Transverse crack. The experimental results confirm that the performance of DDACDN model is still better than Baseline in terms of \textbf{F$_1$} and \textbf{Acc}. Specifically, compared with the Baseline, DDACDN model on the CFD dataset exceeds 0.9\%, 0.8\%, and 0.9\% respectively on the three evaluation metrics. Even if the baseline is already at a very high level of \textbf{F$_1$} and \textbf{Acc} on this dataset, our model can still improve it. Moreover, on the CQU-BPMDD dataset, DDACDN model exceeds the Baseline by 4.2\% and 5.3\% on the first and third metrics, which more fully shows that our method has better cross-data generalization ability than Baseline. 

\subsection{Robustness Analysis}
In this section, we analyze the robustness of our method on $D_t$ (CQU-BPDD). Specifically, Gaussian noise is randomly introduced to each image on $D_t$ test set, destroying some pixels. The noise ratios represent the proportion of damaged pixels in the image, which are 10\%, 20\%, and 30\% respectively. Fig. \ref{fig:11} shows the performance of our method and Baseline at different noise ratios. It can be clearly seen from the observation that our method consistently outperforms Baseline under all noise ratios. The maximum gains of \textbf{F}$_1$ under the four categories are 6.5\%, 5.8\%, 4.9\%, and 11.6\% respectively. As the proportion of noise increases, so does the gain. The top gain is the \textbf{F}$_1$ of Pothole category, which increases from 8\% at 10\% noise ratio to 11.6\% at 30\% noise ratio. All these phenomena show that our method has stronger advantages even in noisy scenarios. Compared with the baseline, YOLOv5, our method enjoys the stronger robustness to noise.
\begin{table}[h]
	\caption{The ablation analysis of each module of our method on CQU-BPDD dataset.}
	\label{tab:4}
	\renewcommand\arraystretch{1.5}
	\centering
	\setlength{\tabcolsep}{2.8mm}{
		\begin{tabular}{ c|  c c c c  }\Xhline{1.2pt}  
			Methods & F$_1^L$ &F$_1^T$&F$_1^A$&F$_1^P$ \\ \hline
			Baseline&81.5\%&87.9\%&93.5\%&62.1\%\\
			Baseline+APAGE&83.6\%&89.7\%&94.2\%&65.6\%\\
			Baseline+APAGE+DA&\textbf{86.9\%}&\textbf{91.5\%}&\textbf{95.6\%}&\textbf{70.0\%}
			\\
			\Xhline{1.2pt}
	\end{tabular}}
\end{table}
\subsection{Ablation Study}
Table \ref{tab:4} shows the results of ablation analysis, where APAGE represents our proposed adaptive patch augmentation and global equalization, $F_1^A$ and $F_1^P$ indicate the \text{F}$_1$ scores of Alligator crack and Pothole, respectively. DA represents our proposed domain adaptation method, including two modules of feature extraction and domain adaptation mining in Section \uppercase\expandafter{\romannumeral3}. Through observation, we can find that the APAGE module and DA module proposed by us have significantly increased all metrics compared with the Baseline. The improvement of two modules in Pothole category is as high as 3.5\% and 7.9\% respectively.
\begin{table}[h]
	\caption{The effect of preprocessing on the performance of our method on CQU-BPDD dataset.}
	\label{tab:5}
	\renewcommand\arraystretch{1.5}
	\centering
	\setlength{\tabcolsep}{3.65mm}{
		\begin{tabular}{ c|  c c c c  }\Xhline{1.2pt}  
			Methods & F$_1^L$ &F$_1^T$&F$_1^A$&F$_1^P$ \\ \hline
			Original Image&84.7\%&88.9\%&93.9\%&62.0\%\\
			CLAHE \cite{109340}&85.4\%&91.2\%&95.0\%&67.7\%\\
			APAGE&\textbf{86.9\%}&\textbf{91.5\%}&\textbf{95.6\%}&\textbf{70.0\%}\\
			\Xhline{1.2pt}
	\end{tabular}}
\end{table}

\begin{table}[h]
		\caption{Comparison between the shared weight and unshared weights. Note that the unshared weights pre-trained with a few samples of the target domain dataset.}
		\label{tab:6}
		\renewcommand\arraystretch{1.5}
		\centering
		\setlength{\tabcolsep}{3.3mm}{
			\begin{tabular}{ c|  c c c c c }\Xhline{1.2pt}  
				Weight & F$_1^L$ &F$_1^T$&F$_1^A$&F$_1^P$&Acc \\ \hline
				Shared&\textbf{86.9\%}&\textbf{91.5\%}&\textbf{95.6\%}&\textbf{70.0\%}&\textbf{82.6\%}\\
				Unshared&84.6\%&89.7\%&94.3\%&64.5\%&80.5\%\\
				\Xhline{1.2pt}
	\end{tabular}
}
\end{table}

Moreover, we also empirically discussed the effect of different preprocessing methods on the performance of our method. Table \ref{tab:5} lists the performance of different preprocessing methods on our model. It can be seen that CLAHE and our method outperform the original unprocessed image in terms of all evaluation metrics. However, compared with CLAHE, our method has an improvement of 3.3\%, 1.8\%, 1.4\%, and 4.4\% on the four metrics respectively. All these results demonstrate show the effectiveness of our method. 

	Finally, we perform ablation experiments on whether the weights of the DDACDN backbone network are shared or not. Specifically, we use the weights pre-trained with a few samples of the target domain dataset to evaluate the network performance under different weights of the backbone. The results are shown in Table \ref{tab:6}. Experiments show that under the same training parameters, crack feature patterns on the source domain cannot transfer well to the target domain with unshared weights backbones, and resulting in insufficient performance compared to sharing weights. Specifically, in the Pothole category, the \text{F}$_1$ score has a decrease of 5.5\% than the shared weight.
	
\subsection{Visualization of Domain Adaptation}
\begin{figure}[h]
	\centering
	\subfigure[Original crack image.]{\includegraphics[scale=0.548]{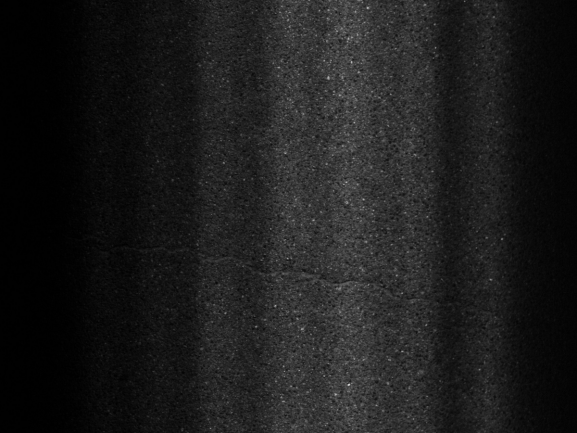}}
	\subfigure[4-th hidden layer]{\includegraphics[scale=0.301]{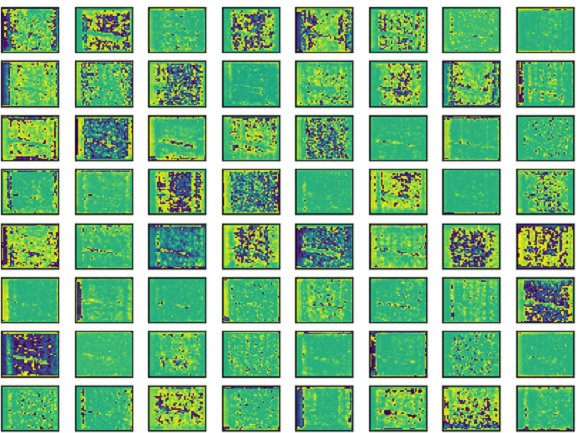}}
	\subfigure[6-th hidden layer]{\includegraphics[scale=0.2999]{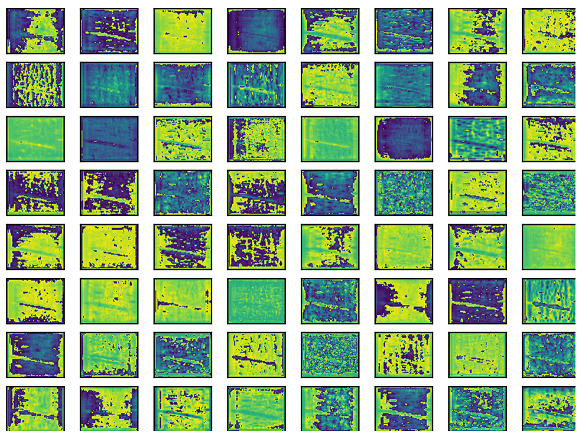}}
	\subfigure[9-th hidden layer]{\includegraphics[scale=0.2998]{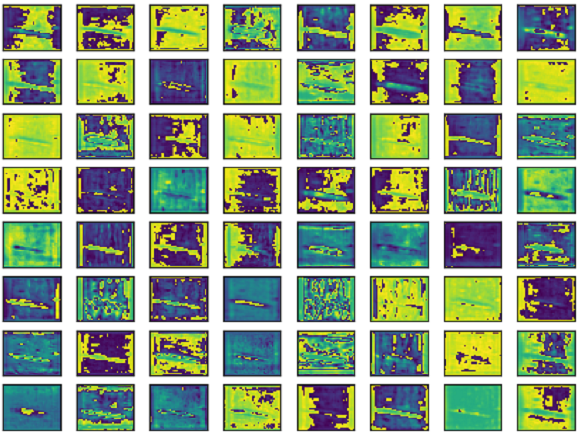}}
	\caption{Visualization of Crack Feature Map in Target Domain, where a, b, c, and d represent the original crack image, feature maps of the 4-th, 6-th, and 9-th hidden layers, respectively. Note that the depth of feature maps is from shallow to deep, and each layer retains 64 images.}
	\label{fig:12}
\end{figure}
	In this section, we focus on visualizing the features that are preserved during domain adaptation. As shown in Fig. \ref{fig:12}, after domain adaptation, the deep network feature map not only focuses on the crack structure, but also pays attention to the background of the crack. Because our purpose is to transfer the crack feature pattern on the source domain to the target domain, so that cracks can also be well detected in the background of the target domain, and it is necessary to transfer the background features to assist the detection while retaining the crack features. In this adaptation mode, if the model is able to detect a certain category of crack in the background of the source domain, it will also be able to identify it in the background of the target domain as well. In summary, the simultaneous adaptation to cracks and background enables the model to detect cracks in different backgrounds.

\section{Conclusion}
In this paper, we have proposed a novel Deep Domain Adaptation-based Crack Detection Network (DDACDN) for multi-category crack detection, which takes advantage of the knowledge of the source domain to help detect the target domain without annotated information. DDACDN adopts a Multi-scale Domain Adaptation (MDA) strategy, which uses MK-MMD to update domain loss on the feature space of three scales in two domains. Then, it takes advantage of the features extracted from the source domain and the target domain to construct an intermediate domain and use it to train the network. 
Moreover, in order to effectively suppress the negative effects of illumination, we proposed a novel and a simple preprocessing method called APAGE to enhance the image through patch augmentation and global equalization. Four datasets are used to evaluate the DDACDN model. More specifically, A large-scale Bituminous Pavement Multi-label Disease Dataset named CQU-BPMDD is constructed to evaluate the cross-data generalization ability of our work. Experimental results demonstrate that DDACDN has superior detection performance to some state-of-the-art CNN methods on CQU-BPDD dataset. Furthermore, excellent performance is achieved on other datasets, such as CQU-BPMDD and CFD. In addition, DDACDN not only provides accurate pavement disease location information in real scenes, which is convenient for maintenance, but also provides a domain adaptive idea for solving cross-domain crack detection.

In the future, for the problem of data annotation, we will explore the method based on active learning to solve the detection problems of difficult samples and noisy samples. In addition, we will make sufficient analysis on the quantitative evaluation of different crack detection methods, and further investigate methods to improve the detection performance of multi-scale crack objects.


\ifCLASSOPTIONcaptionsoff
  \newpage
\fi



%
\bibliographystyle{IEEEtran}

\bibliography{IEEE_Reference}




\end{document}